\documentclass[10pt,twocolumn,letterpaper]{article}

\usepackage[pagenumbers]{cvpr} 

\usepackage[dvipsnames]{xcolor}

\def\*#1{\mathbf{#1}}
\DeclareMathOperator*{\argmin}{arg\,min}

\usepackage{xcolor}
\definecolor{tabblue}{rgb}{0.54, 0.81, 0.94}
\usepackage{listings}
\usepackage{color, colortbl}

\usepackage{algorithm}
\usepackage{algpseudocode}

\newcommand{\RR}{\mathbb{R}}
\newcommand{\Ff}{\mathcal{F}}
\usepackage[accsupp]{axessibility}

\definecolor{cvprblue}{rgb}{0.21,0.49,0.74}
\usepackage[pagebackref,breaklinks,colorlinks,citecolor=cvprblue]{hyperref}

\def\paperID{12673} 
\def\confName{CVPR}
\def\confYear{2024}

\title{Memory-Scalable and Simplified Functional Map Learning}

\author{Robin Magnet\\
LIX, \'Ecole Polytechnique, IP Paris\\
{\tt\small rmagnet@lix.polytechnique.fr}
\and
Maks Ovsjanikov\\
LIX, \'Ecole Polytechnique, IP Paris\\
{\tt\small maks@lix.polytechnique.fr}
}

\begin{document}
\maketitle
\begin{abstract}
Deep functional maps have emerged in recent years as a prominent learning-based framework for non-rigid shape matching problems. While early methods in this domain only focused on learning in the functional domain, the latest techniques have demonstrated that by promoting consistency between functional and pointwise maps leads to significant improvements in accuracy. Unfortunately, existing approaches rely heavily on the computation of large dense matrices arising from soft pointwise maps, which compromises their efficiency and scalability. To address this limitation, we introduce a novel memory-scalable and efficient functional map learning pipeline. By leveraging the specific structure of functional maps, we offer the possibility to achieve identical results without ever storing the pointwise map in memory. Furthermore, based on the same approach, we present a differentiable map refinement layer adapted from an existing axiomatic refinement algorithm. Unlike many functional map learning methods, which use this algorithm at a post-processing step, ours can be easily used at train time, enabling to enforce consistency between the refined and initial versions of the map. Our resulting approach is both simpler, more efficient and more numerically stable, by avoiding differentiation through a linear system, while achieving close to state-of-the-art results in challenging scenarios.
\end{abstract}    
\section{Introduction}\label{sec:Introcution}

Automatically computing dense correspondences between non-rigid shapes is a classical problem in computer vision, forming the foundation of various downstream applications like shape registration~\cite{bogoDynamicFAUSTRegistering2017}, deformation~\cite{sorkineAsRigidAsPossibleSurfaceModeling, dengSurveyNonRigid3D2022}, and analysis~\cite{sahilliogluRecentAdvancesShape2020}.
A popular approach to tackle this problem involves the functional map pipeline~\cite{ovsjanikovFunctionalMapsFlexible2012}, which represents correspondences as linear operators between functional spaces derived from the intrinsic Laplacian~\cite{meyerDiscreteDifferentialGeometryOperators2003} on each shape.  Numerous early methods~\cite{nognengImprovedFunctionalMappings2018, renContinuousOrientationpreservingCorrespondences2019, nognengInformativeDescriptorPreservation2017} have leveraged this framework using handcrafted descriptors to generate functional maps, which can lack fine detail. Many algorithms~\cite{melziZoomOutSpectralUpsampling2019, renDiscreteOptimizationShape2021, ezuzDeblurringDenoisingMaps2017, magnetSmoothNonRigidShape2022} have therefore successfully been developed in order to refine such imprecise maps into high quality dense correspondences.

\begin{figure}
    \centering
    \includegraphics[width=.95\linewidth]{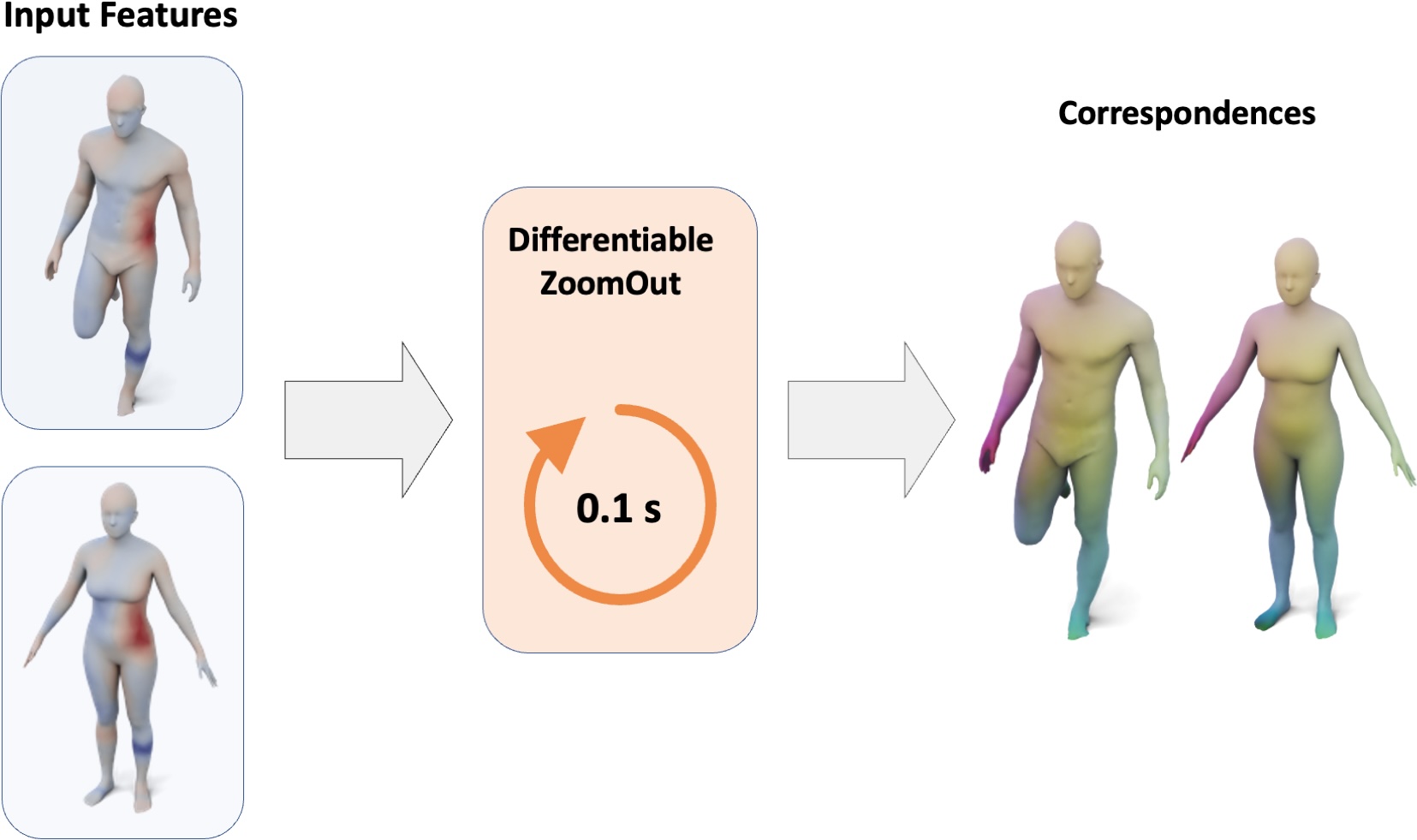}
    \caption{Our method takes a set of point features as input, which can be learned, and uses a differentiable version of the ZoomOut algorithm to produce correspondences. Due to its light memory cost, it can be used while training a network, or when running the network on very dense meshes.}
    \label{fig:Teaser}\vspace{-12pt}
\end{figure}

Building upon pioneering efforts by \cite{donatiDeepGeometricFunctional2020}, recent advancements~\cite{sharpDiffusionNetDiscretizationAgnostic2022a, eisenbergerDeepShellsUnsupervised2020, sunSpatiallySpectrallyConsistent2023} have successfully explored the possibility of \textit{learning} descriptors directly from data for subsequent functional map computations, adapting the original pipeline introduced by \cite{ovsjanikovFunctionalMapsFlexible2012,roufosseUnsupervisedDeepLearning2019}. Notably, the most recent developments in this area have observed that promoting functional maps to be  ``proper'' (i.e., functional maps arising from pointwise ones) can lead to significant improvement in accuracy. The concept of  ``proper'' functional maps was introduced in the optimization setting ~\cite{renDiscreteOptimizationShape2021} and then quickly adopted within the learning context. Specifically,  recent deep functional map methods have constructed dual-branch networks~\cite{attaikiUnderstandingImprovingFeatures2023, liLearningMultiresolutionFunctional2022, caoUnsupervisedLearningRobust2023, sunSpatiallySpectrallyConsistent2023} that enforce the connection between pointwise and functional maps and that have demonstrated impressive performance across multiple datasets. Interestingly, these studies highlighted the necessity of retaining the original functional map branch~\cite{sunSpatiallySpectrallyConsistent2023} to achieve optimal performance, despite its inherent instability when differentiating through the linear system solver~\cite{efroniSpectralTeacherSpatial2022}.

In all these works, however, the ``properness' of functional maps  is enforced by first computing a soft point-to-point map which is then converted to a functional map using matrix multiplication. This heavily limits the scalability of these approaches, as the dense pointwise map has to be stored in memory, which scales quadratically with the number of vertices. While common shape matching benchmarks only use meshes with low number of vertices, using these methods on real meshes is a serious challenge.

To address this limitation, we propose an approach that can compute the functional map associated with the soft p2p map, without \textit{ever} storing the dense matrix in memory. Key to our approach is the fact that the proper functional map is defined as a matrix product between the soft pointwise map and the Laplacian basis~\cite{ovsjanikovFunctionalMapsFlexible2012,renDiscreteOptimizationShape2021}. By exploiting this structure and GPU acceleration~\cite{Keops2021}, we show that such matrix product can be computed directly without the necessity of storing the pointwise map, thus significantly improving both the speed and scalability of related approaches.

Our work additionally demonstrates the feasibility of discarding the original functional branch while preserving result quality. Our approach involves the transformation of a widely adopted map refinement algorithm~\cite{melziZoomOutSpectralUpsampling2019}, originally implemented on CPU, into a differentiable and memory-efficient GPU version using a similar pointwise map computation. Utilizing this refined map allows us to impose constraints on the structure of the learned functional map through a form of self-supervision. This, in turn, replaces the need for a consistency loss with the traditional functional map branch as in~\cite{caoUnsupervisedLearningRobust2023, sunSpatiallySpectrallyConsistent2023}, providing a novel simple and efficient solution for maintaining result quality in the absence of the original functional branch. Overall, our contributions can be summarized as follows:
\begin{itemize}
    \item We propose efficient GPU implementation of differentiable pointwise map or functional map learning with minimal space complexity and numerical stability.
    \item We use a novel GPU adapted refinement algorithm at train time to provide self-supervision to the network.
    \item We introduce the first single-branch network for functional map learning without differentiating through a linear system solver.
\end{itemize}
\section{Related Works} \label{sec:Related Works}

Shape matching and in particular functional map correspondence computation is a very wide and established a field of research. We here only review the works the closest to our work, and refer the interested reader to~\cite{ovsjanikovComputingProcessingCorrespondences2017, sahilliogluRecentAdvancesShape2020} for an in-depth description of related works.

\paragraph{Functional Maps} Our work is built upon the functional map framework, originally developed in ~\cite{ovsjanikovFunctionalMapsFlexible2012} and later extended in various ways~\cite{nognengImprovedFunctionalMappings2018, renContinuousOrientationpreservingCorrespondences2019, melziZoomOutSpectralUpsampling2019, renDiscreteOptimizationShape2021, donatiComplexFunctionalMaps2022a}, an overview being provided in~\cite{ovsjanikovComputingProcessingCorrespondences2017}. This approach encodes correspondences between shapes as small sized matrices independently of the original number of vertices, offering an efficient way to compute maps. This then allows to efficiently enforce constraints on the correspondences such as bijectivity or area preservation using simple linear algebra. The most effective functional map algorithms are map refinement algorithms~\cite{melziZoomOutSpectralUpsampling2019, renDiscreteOptimizationShape2021, magnetSmoothNonRigidShape2022, eisenbergerSmoothShellsMultiScale2020}, which take initial correspondences as input and iteratively refine them. While highly robust, obtaining initialization without landmarks often relies on the use of handcrafted descriptors such as HKS~\cite{bronsteinScaleinvariantHeatKernel2010}, WKS~\cite{aubryWaveKernelSignature2011} or SHOT~\cite{tombariUniqueSignaturesHistograms2010}.

\paragraph{Deep Functional Maps} A more recent line of research focus on learning descriptor functions directly from the surface itself. Originating with FMNet~\cite{litanyDeepFunctionalMaps2017, roufosseUnsupervisedDeepLearning2019} and further developed in~\cite{donatiDeepGeometricFunctional2020, sharpDiffusionNetDiscretizationAgnostic2022a}, these approaches typically take handcrafted descriptors as inputs and yield refined descriptor functions. These functions are then used in a standard functional map pipeline~\cite{ovsjanikovFunctionalMapsFlexible2012}, and are usually post-processed at test time using off-the-shelf map refinement algorithms~\cite{melziZoomOutSpectralUpsampling2019, vestnerProductManifoldFilter2017, ezuzDeblurringDenoisingMaps2017, renContinuousOrientationpreservingCorrespondences2019}. Using modern feature extractors for surfaces and point cloud~\cite{sharpDiffusionNetDiscretizationAgnostic2022a, thomasKPConvFlexibleDeformable2019}, these works obtained impressive results despite the unstable differentiation through a linear system solver~\cite{efroniSpectralTeacherSpatial2022}.
While these initial approaches primarily focused on supervised learning, contemporary research in functional map learning emphasizes unsupervised learning of correspondences~\cite{sunSpatiallySpectrallyConsistent2023,caoUnsupervisedDeepMultishape2022,caoUnsupervisedLearningRobust2023, liLearningMultiresolutionFunctional2022}. This is achieved using functional map priors, that is, explicitly promoting structural properties on the learned functional map such as orthogonality - which corresponds to area preservation in the spatial domain.
Recent advancements~\cite{renDiscreteOptimizationShape2021} have highlighted the importance of using extra structural constraint in the form of ``proper'' functional maps, that are functional maps obtained from pointwise correspondences, a guarantee not provided in the original pipeline~\cite{ovsjanikovFunctionalMapsFlexible2012} or learning-based approaches~\cite{donatiDeepGeometricFunctional2020}.
This led to the development of methods computing a second functional map at train-time using soft correspondences, resulting in dual-branches networks~\cite{attaikiUnderstandingImprovingFeatures2023, sunSpatiallySpectrallyConsistent2023, caoUnsupervisedLearningRobust2023, caoUnsupervisedDeepMultishape2022}. These approaches were however recognized~\cite{sunSpatiallySpectrallyConsistent2023} as unable to scale to large meshes, due to large dense matrix computations, and had to use mesh resampling to avoid memory and speed issues.

\paragraph{Differentiable Refinement} In a context also aligned with our work, it was noted in~\cite{liLearningMultiresolutionFunctional2022} that proper functional maps were guaranteed by many map refinement algorithms~\cite{melziZoomOutSpectralUpsampling2019, renDiscreteOptimizationShape2021, magnetSmoothNonRigidShape2022}. Subsequently, this refinement was partially integrated into a network as a differentiable post-processing step for the initially learned functional map. However, the design from~\cite{liLearningMultiresolutionFunctional2022} still relies on the original linear system solver, and their adaptation of~\cite{melziZoomOutSpectralUpsampling2019} was only partial.
This partial adaptation was necessitated by the potential memory overflow resulting from numerous dense map computations. Additionally, the output functional map was only a weighted sum of proper functional maps, thus lacking a guarantee of being proper itself.
\begin{figure*}
  \centering
  \includegraphics[width=.9\linewidth]{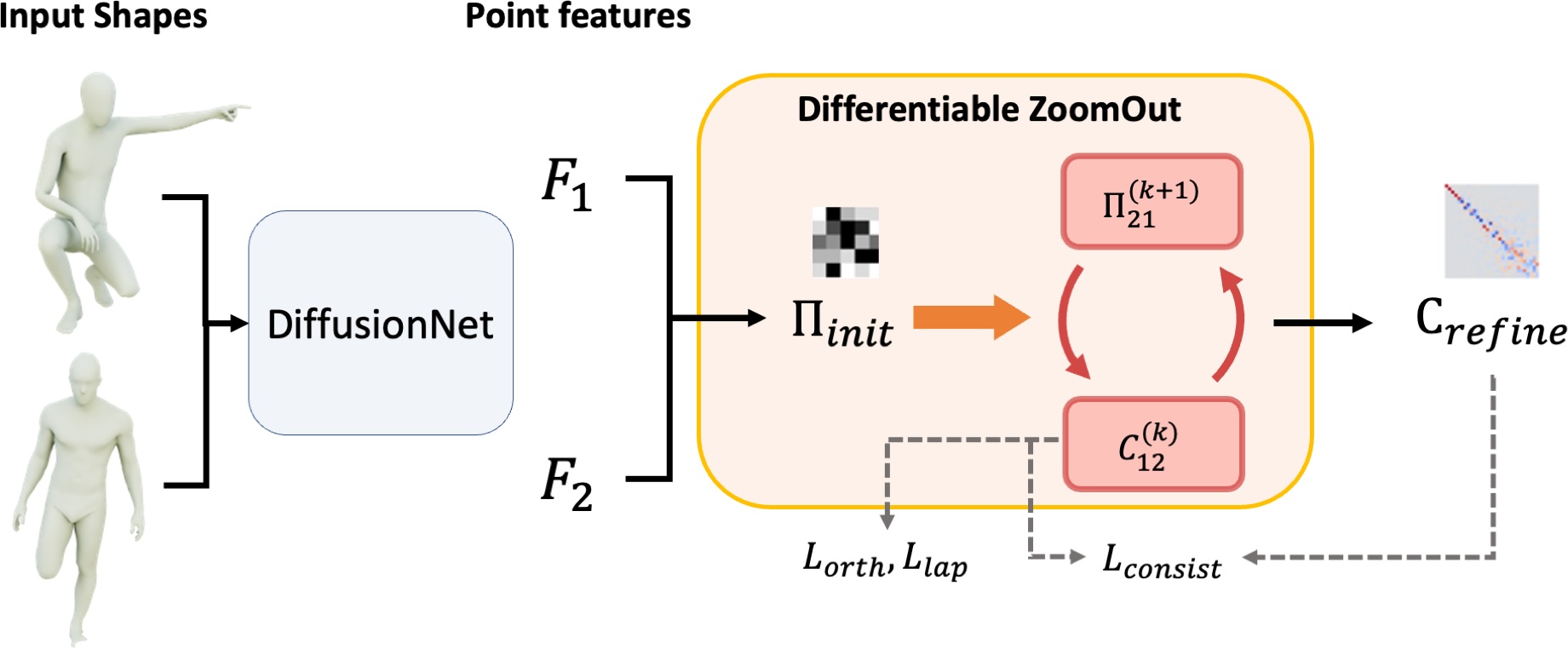}
    \caption{Our pipeline takes as input two shapes and use a feature extractor network to obtain pointwise features. These features are used to compute an initial pointwise map and then fed to our Differentiable ZoomOut block. All the pointwise maps $\Pi$ are our scalable dense maps, which are memory efficient.}
    \label{fig:pipeline}

\end{figure*}
\section{Background \& Motivation}
\label{sec:Notations and background}

Our method builds upon the functional map framework~\cite{ovsjanikovFunctionalMapsFlexible2012}, and in particular of its recent development, using learning-based descriptors inspired by GeoFMaps~\cite{donatiDeepGeometricFunctional2020}. Before describing our approach in \Cref{sec:Method}, we provide an overview of the foundation of this pipeline. Interested readers are encouraged to explore numerous related works~\cite{attaikiUnderstandingImprovingFeatures2023, sunSpatiallySpectrallyConsistent2023, liLearningMultiresolutionFunctional2022, caoUnsupervisedDeepMultishape2022, caoUnsupervisedLearningRobust2023, ovsjanikovComputingProcessingCorrespondences2017} for additional insights into various adaptations and nuances of this framework.

\paragraph{Notations} We will suppose to be given two shapes $S_1$ and $S_2$ with respectively $n_1$ and $n_2$ vertices. For each shape $S_i$, we compute its intrinsic Laplacian~\cite{meyerDiscreteDifferentialGeometryOperators2003}, and store its eigenfunctions as columns of a matrix $\Phi_i\in\RR^{n_i\times K}$. We denote $\Phi_i^\dagger=\Phi_i^\top A_i$ its pseudo-inverse, with $A_i$ being the diagonal vertex-area matrix.  Given any matrix $B$, we denote $[B]_i$ the vector consisting of the $i$-th line of $B$.

\paragraph{Deep Functional Maps} The standard deep functional map pipeline~\cite{donatiDeepGeometricFunctional2020} takes 2 shapes $S_1$ and $S_2$ as input, and use a feature extractor network $\Ff_\theta$ to generate $p$ descriptor functions on each shape, stored as columns of matrices $F_i=\Ff_\theta(S_i)\in\RR^{n_i\times p}$. Following the standard functional map pipeline~\cite{ovsjanikovFunctionalMapsFlexible2012} these descriptors are first projected into the Laplacian basis $\*A_i=\Phi_i^\dagger F_i\in\RR^{K\times p}$ and a functional map is obtained by solving the linear system:
\begin{equation}\label{eq:C linear}
    \argmin_\*C \| \*C \*A_1 - \*A_2 \|_2^2.
\end{equation}
This linear system is further usually regularized using an extra Laplacian term~\cite{donatiDeepGeometricFunctional2020, renStructuredRegularizationFunctional2019}.
During training, losses are then imposed on the computed functional map $\*C\big(\Ff_\theta(S_1), \Ff_\theta(S_2)\big)$. At test time, a pointwise map can be recovered from the map $\*C$ and the eigenfunctions $\Phi_i$ using nearest neighbor search~\cite{ovsjanikovFunctionalMapsFlexible2012, paiFastSinkhornFilters2021}.

\paragraph{Two branches networks} Recent works in functional map literature~\cite{renDiscreteOptimizationShape2021} have highlighted the positive effects of using \emph{proper} functional maps. A functional map is proper if it arises from \textit{some} underlying pointwise map. Specifically, a proper functional map is \emph{defined} as the pull-back of a pointwise map $T:S_2\to S_1$:
\begin{equation}\label{eq:C pull back}
    \*C = \Phi_2^\dagger \Pi \Phi_1
\end{equation}
where $\Pi\in\{0,1\}^{n_2\times n_1}$ is the matrix representation of the map $T$. Several works~\cite{attaikiUnderstandingImprovingFeatures2023, sunSpatiallySpectrallyConsistent2023, caoUnsupervisedLearningRobust2023} adopt a differentiable approach to compute $\Pi$ before deriving $\*C_{\text{proper}}$ using \Cref{eq:C pull back}.
Typically, the map $\Pi$ is computed from the features $F_1$ and $F_2$ using a Gaussian kernel:
\begin{equation}\label{eq:Kernel map}
    \Pi_{ij} = \frac{\exp(\delta_{ij})}{\sum_k\exp(\delta_{ik})}
\end{equation}
with $\delta_{ij}=-\frac{1}{2\sigma^2}\|[F_2]_i - [F_1]_j\|^2$ the distance between rows of the feature matrices, where $\sigma$ a temperature - or blur parameter.
For training purposes, a consistency loss between $\*C$, obtained with \cref{eq:C linear}, and $\*C_{\text{proper}}$, derived using  \cref{eq:C pull back,eq:Kernel map}, is employed. This approach is taken \textit{in addition} to the standard orthogonality or bijectivity losses presented in~\cite{donatiDeepGeometricFunctional2020}.

\paragraph{ZoomOut} A popular map refinement algorithm named ZoomOut~\cite{melziZoomOutSpectralUpsampling2019} has often been used to obtain high-quality correspondences from low quality initial functional maps such as those obtained from learning pipelines. ZoomOut iteratively computes functional maps using \cref{eq:C pull back} and pointwise map using nearest neighbor search between the rows of $\Phi_1 \*C^T$ and $\Phi_2$. Note that due to its iterative nature, ZoomOut is \textit{guaranteed} to produce proper functional maps. A recent \textit{approximation}~\cite{magnetScalableEfficientFunctional2023} made the algorithm scalable to dense meshes on CPU, but however relies on sampling, a longer pre-processing and a final slow conversion from the samples back to the full shapes.

\paragraph{Drawbacks and motivation}
Despite achieving high quality results on shape matching benchmarks, the modern two-branches approaches presented above suffer from three notable drawbacks.
Firstly, computing $\Pi$ using~\cref{eq:Kernel map} involves storing and differentiating through a dense $n_2\times n_1$ matrix, making the method scale poorly in terms of memory. In particular, because of the linear system used in the other branch, features are required to be of high dimension (usually 128 or 256) to ensure invertibility of the feature matrix, thus heavily slowing down computations.
Secondly, a naive implementation of \cref{eq:Kernel map} can result in underflows in the forward or backward pass for low values of $\sigma$.
Thirdly, as remarked in some previous works \cite{efroniSpectralTeacherSpatial2022} despite its necessity for achieving satisfactory results, the original functional map branch from~\cite{donatiDeepGeometricFunctional2020} poses a risk of instability due to differentiation through the linear system solver.

In this work, we seek to address these challenges by establishing soft point-wise maps as a stable and memory-scalable option to learn functional maps, without approximations such as those presented in~\cite{magnetScalableEfficientFunctional2023}. A second goal lies in trying to completely remove the spectral branch from the learning procedure. 
The necessity of the spectral branch suggested in~\cite{sunSpatiallySpectrallyConsistent2023} hints that properness might not be a sufficient constraint alone for efficient learning of correspondences. To overcome this challenge, we further refine the structural constraints by introducing the expectation that the functional map aligns with its refined version, produced by~\cite{melziZoomOutSpectralUpsampling2019}. This leads to the first deep functional map method that completely avoids solving a linear system inside the network, enables unsupervised training, is scalable, efficient and leads to high quality results.

\section{Method}
\label{sec:Method}

In this section, we introduce our scalable approach to proper functional maps, which we then apply to design a novel GPU based differentiable version of the ZoomOut~\cite{melziZoomOutSpectralUpsampling2019} algorithm. Finally, using these two elements, we introduce our new single branch network for functional map learning without a linear system solver.

\subsection{Scalable Dense Maps}

The key observation to this work is that all dense pointwise maps computed in deep functional map  pipelines~\cite{liLearningMultiresolutionFunctional2022, sunSpatiallySpectrallyConsistent2023, attaikiDPFMDeepPartial2021, caoUnsupervisedLearningRobust2023} are used \textit{exclusively} to compute functional maps using \Cref{eq:C pull back}. In particular, they are invariably found in a matrix product of the form $\Pi \Phi_1$. Previous work did not seek to exploit this fact, and instead computed the complete dense matrix $\Pi$ separately before performing the matrix product.
In contrast, we argue it is possible to compute the result of this inner product without ever computing any dense $n_2\times n_1$ matrix.

\begin{figure}
    \centering
    \includegraphics[width=.8\linewidth]{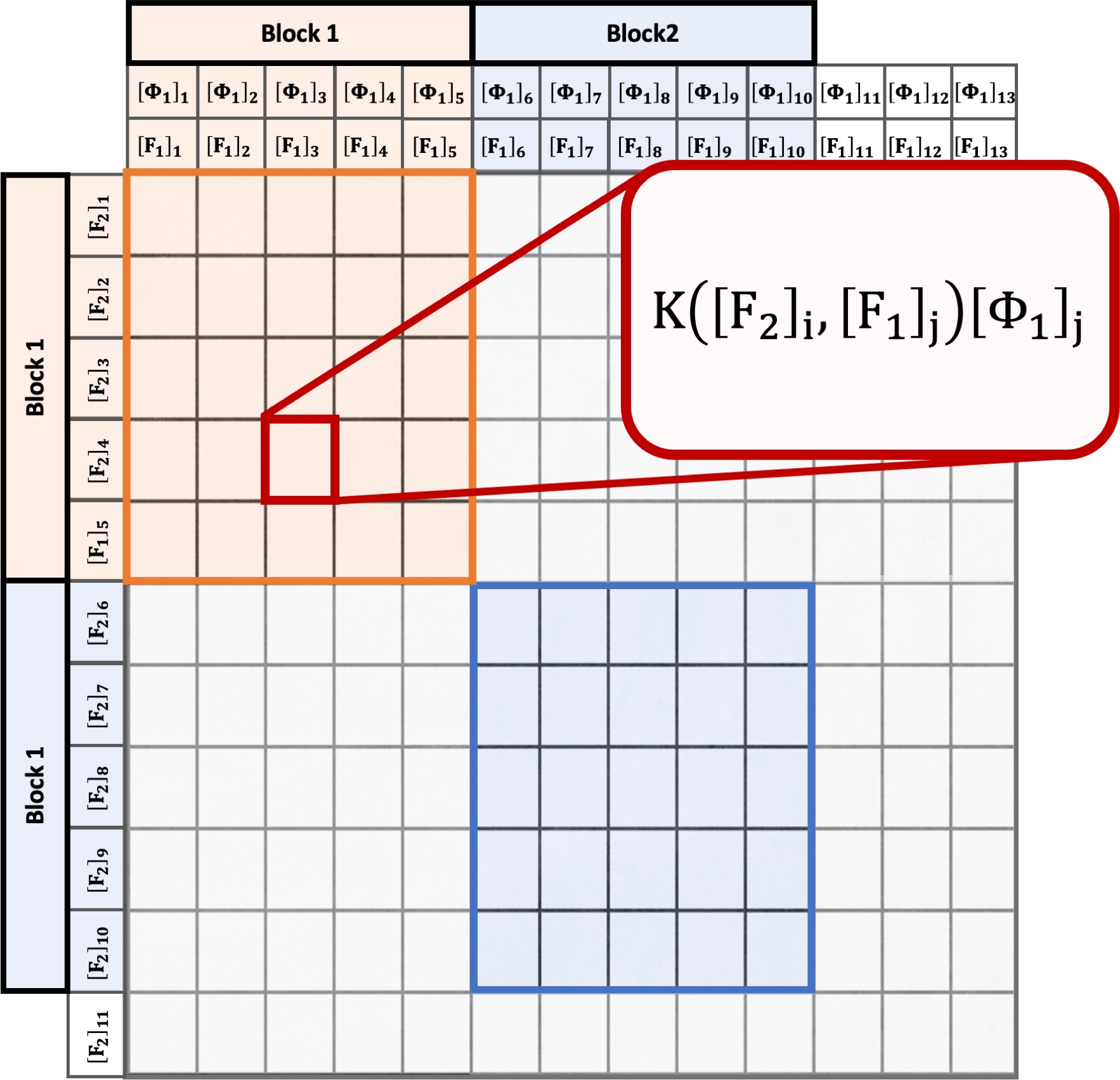}
    \caption{Our scalable dense maps relies on the underlying structure of \cref{eq:Pull-back with K}, where the sum is computed for each contiguous memory block highlighted in the image. The entries are evaluated on the fly while performing summation, and results from each block are then accumulated to obtain the final per-rows values. The implementation is provided by the Keops package~\cite{Keops2021}.}
    \label{fig:Keops} \vspace{-12pt}
\end{figure}

Observe first that we can explicitly write the $i$-th line of $\Pi\Phi_1$, using \Cref{eq:Kernel map} as:
\begin{align}\label{eq:Pull-back sum}
    [\Pi\Phi_1]_{i} &= \sum_{j=1}^{n_1} \frac{\exp(\delta_{ij})}{\sum_k\exp(\delta_{ik})} [\Phi_1]_{j} \\
    &= L_i^{-1} \sum_{j=1}^{n_1} K\big([F_2]_i, [F_1]_j\big) [\Phi_1]_{j} \label{eq:Pull-back with K}.
\end{align}
where $K$ is an RBF Kernel, and $L_i$ the row normalization. By rewriting the proper functional map definition in this kernel form, we can now leverage existing methods for heavily scalable and fast GPU computation with kernels~\cite{rudiFALKONOptimalLarge2017, meantiKernelMethodsRoof2020, Keops2021}.
These methods rely on, in particular, the fact that the entry $(i,j)$ of the Kernel matrix $K=\big(\exp(\delta_{ij})\big)_{ij}$ \emph{only} depends on the vectors $[F_2]_i$ and $[F_1]_j$. This allows to compute the sum in \Cref{eq:Pull-back with K} in a block-wise manner, where the values of $K$ are computed during summation. This is highlighted in \Cref{fig:Keops}, where we represent the dense matrix on which summation in applied in \Cref{eq:Pull-back with K}. The per-row sum can then be computed first for each contiguous memory block before summing all the outputs to obtain the value of $\Pi\Phi_1$.

In practice, we rely on the Keops library~\cite{Keops2021}, which applies such operations on very large dense matrices whose entries can be described by mathematical formulas applied to the inputs. Keops uses symbolic matrices, and computes reduction on-the-fly using per-block operations for fast computation without ever fitting the dense matrix in memory.

Note that the normalization $L_i$ can additionally be handled using efficient stabilized logsumexp reductions and incorporated into the Kernel K to avoid underflow or overflow in the exponential. Furthermore, the gradient of $\Pi \Phi_1$ with respect to $F_1$ and $F_2$ can be computed using a similar trick~\cite{Keops2021}.

At test time, a vertex-to-vertex map can be extracted from $\Pi$ by looking for the indices of the per-row maximal value, which is equivalent to running nearest neighbor search between the rows of $F_1$ and the rows of $F_2$. This can again be run efficiently on GPU without computing the dense distance matrix, using GPU-based nearest neighbor implementations~\cite{Keops2021, johnson2019billion}.

Ultimately, our dense pointwise map only stores values for $F_1$ and $F_2$ as well as a the type of Kernel we use, and has therefore a linear memory cost.

\subsection{Differentiable ZoomOut}
\label{subsec:Differentiable ZoomOut}
As mentioned in \Cref{sec:Notations and background}, the ZoomOut algorithm~\cite{melziZoomOutSpectralUpsampling2019} iteratively performs pointwise map computations using nearest neighbor queries between rows of $\Phi_1 \*C_{12}^\top$ and of $\Phi_2$, and functional map computations using~\cref{eq:C pull back}, while increasing the size $K$ of the spectral basis. By replacing the nearest neighbor queries by differentiable soft maps that we store using our scalable versions, we introduce Differentiable ZoomOut, a fast and fully differentiable block, with negligible memory cost. The algorithm is presented in detail in the supplementary material.

Since ZoomOut acts as a powerful map refinement algorithm, we would like to enforce consistency between the output and input functional maps of the ZoomOut algorithm in order to help training. We expect such a loss to provide meaningful guidance to the features.

However, we note that the output functional map $\*C_{\text{refine}}$ has a larger size than the initial map $\*C_{\text{init}}$. This is due to ZoomOut using an increasing size of spectral basis.
However, given a proper functional map of size $K_2\times K_1$ associated to a pointwise map $\Pi$, the principal submatrix composed of its first $K_2'$ rows and $K_1'$ column from the proper functional map of size $K_2'\times K_1'$ associated to the same map $\Pi$. This stems from the definition of proper functional maps~\cite{ovsjanikovFunctionalMapsFlexible2012}, and we refer to the supplementary for details on this aspect.
Therefore, our new consistency loss only uses a principal submatrix of the refined functional map:
\begin{equation}\label{eq:loss consist}
    L_{\text{consist}}( \*C_{\text{init}}, \*C_{\text{refine}}) = \| \*C_{\text{init}}-[\*C_{\text{refine}}]_{1:K_\text{init}, 1:K_\text{init}}\|_2^2
\end{equation}
where $K_\text{init}$ is the size of the input functional map.


\subsection{Overall Pipeline and Implementation}
\label{subsec:Pipeline}
We would first like to highlight that our scalable dense maps can be used in any existing functional map base model using dense pointwise maps, with no impact on the results. Furthermore, we present a novel single-branch network for functional map prediction which exploits the structural properties of proper functional maps and does not require solving or differentiating through a linear system. We therefore present separate implementations first for our scalable dense maps and differentiable ZoomOut at \url{https://github.com/RobinMagnet/ScalableDenseMaps}, and of our entire pipeline at \url{https://github.com/RobinMagnet/SimplifiedFmapsLearning}. 

As shown in \Cref{fig:pipeline}, our algorithm first extracts features from surfaces $S_1$ and $S_2$ using DiffusionNet~\cite{sharpDiffusionNetDiscretizationAgnostic2022a}. This produces matrices of features $F_1\in\RR^{n_1\times p}$ and $F_2\in\RR^{n_2\times p}$. Importantly, we select $p=32$ instead of the 128 or 256 Features produced by standard pipelines~\cite{liLearningMultiresolutionFunctional2022, caoUnsupervisedDeepMultishape2022, sunSpatiallySpectrallyConsistent2023, caoUnsupervisedLearningRobust2023, attaikiUnderstandingImprovingFeatures2023}, as our approach does not require invertibility of a linear system obtained from the learned features.

An initial soft pointwise map $\Pi_{\text{init}}$ is produced from the features using \Cref{eq:Kernel map}, and then fed into our Differentiable ZoomOut algorithm presented in \Cref{subsec:Differentiable ZoomOut} where we perform $10$ iteration with a spectral step size of $10$ starting with $K_{\text{init}}=30$. This results in a refined map $\*C_{\text{final}}$ of size $K_{\text{final}}=130$. This whole process uses a blur parameter $\sigma=10^{-2}$, which is much lower than previous implementations~\cite{shiUnsupervisedDeepShape2020, liLearningMultiresolutionFunctional2022, sunSpatiallySpectrallyConsistent2023}.


Our unsupervised training loss consists in 3 terms. First, an orthogonality constraint $L_{\text{orth}}(\*C_{\text{init}})=\|\*C_{\text{init}}^\top \*C_{\text{init}} - I\|_2^2$ is applied to the initial functional map, with a weight of $1$.
The ZoomOut consistency loss from~\Cref{eq:loss consist} is applied with an initial weight of $10^{-4}$, gradually increased to $10^{-1}$. This term is therefore ignored during the first epochs until decent initialization has been found. We refer to the supplementary for more details on this aspect.
We finally regularize the result using a Laplacian commutativity term as presented in~\cite{renStructuredRegularizationFunctional2019, caoUnsupervisedDeepMultishape2022}, which is a residual from the spectral branch we discarded. This final term receives a weight of $10^2$.
Eventually, we train our network using ADAM optimizer~\cite{kingmaAdamOptimizer2015} with an initial learning rate of $10^{-3}$. We refer the reader to the supplementary for some more precise details on the implementation.

\begin{table*}
\centering
{\def\arraystretch{1}\tabcolsep=1em

\begin{tabular}{cccccccccc}\toprule[0.8pt]
Train & \multicolumn{3}{c}{\textbf{F}} & \multicolumn{3}{c}{\textbf{S}} & \multicolumn{3}{c}{\textbf{F+S}} \\ \cmidrule(lr){2-4} \cmidrule(lr){5-7} \cmidrule(lr){8-10}
Test & F & S & S19 & F & S & S19 & F & S & S19  \\ \midrule[0.8pt]

BCICP~\cite{renContinuousOrientationpreservingCorrespondences2019} & 6.1 &- &- & - &11.& -& - &- &-\\
ZoomOut~\cite{melziZoomOutSpectralUpsampling2019} & 6.1 &- &- & - &7.5& - &-& -& -\\
SmoothShells~\cite{eisenbergerSmoothShellsMultiScale2020} & 2.5 &- &- &- &4.7 &- &-& -& -\\ 
DiscreteOp~\cite{renDiscreteOptimizationShape2021} &5.6 & - &- & - & 13.1 &- & - & - & -\\\midrule[0.8pt]
GeomFmaps~\cite{donatiDeepGeometricFunctional2020} & 3.5 & 4.8 & 8.5 & 4.0 & 4.3 & 11.2 & 3.5 & 4.4  & 7.1 \\
Deep Shells~\cite{eisenbergerDeepShellsUnsupervised2020} & 1.7 & 5.4 & 27.4 & 2.7 & 2.5 & 23.4 & 1.6 &  2.4 & 21.1 \\
NeuroMorph~\cite{eisenbergerNeuroMorphUnsupervisedShape} & 8.5 & 28.5 & 26.3 & 18.2 & 29.9 & 27.6 &  9.1 & 27.3 & 25.3 \\
DUO-FMNet~\cite{donatiDeepOrientationawareFunctional2022} & 2.5 &  4.2 & 6.4 & 2.7 & 2.6 & 8.4 & 2.5 & 4.3 & 6.4 \\
UDMSM~\cite{caoUnsupervisedDeepMultishape2022} & \textbf{1.5} & 7.3 & 21.5 & 8.6 & 2.0 & 30.7 & 1.7 & 3.2 & 17.8 \\
ULRSSM~\cite{caoUnsupervisedLearningRobust2023} &  1.6 & 6.4 & 14.5 & 4.5 & \textbf{1.8} & 18.5 & \textbf{1.5} & \textbf{2.0} & 7.9  \\
ULRSSM (w/ fine-tune)~\cite{caoUnsupervisedLearningRobust2023} & 1.6 & \textbf{2.2} & 5.7 & 1.6 & 1.9 & 6.7 & 1.6 & 2.1 & 4.6  \\
AttentiveFMaps~\cite{liLearningMultiresolutionFunctional2022} & 1.9 & 2.6 & 5.8 &  1.9 & 2.1 & 8.1 & 1.9 & 2.3 & 6.3 \\
ConsistentFMaps~\cite{sunSpatiallySpectrallyConsistent2023} & 2.3 & 2.6 & \textbf{3.8} & 2.5 & 2.4 & \textbf{4.5} & 2.2 & 2.3 &  4.3 \\ \midrule[0,5pt]
Ours & 1.9 & 2.4 & 4.2 & 1.9 & 2.4 & 6.9 & 1.9 & 2.3 & \textbf{3.6}\\
\bottomrule[0.8pt]
\end{tabular}}
\caption{Mean geodesic errors ($\times 100$) when training and testing on the Faust, Scape and Shrec19 datasets. Best result is shown in bold.} \label{tab:res:all humans} \vspace{-12pt}
\end{table*}
\subsection{Properties of learned features}

An interesting aspect of the two-branches networks~\cite{attaikiUnderstandingImprovingFeatures2023, sunSpatiallySpectrallyConsistent2023, caoUnsupervisedLearningRobust2023} is that each branch offers a different interpretation of the learned features. On the one hand, the standard functional map branch~\cite{ovsjanikovFunctionalMapsFlexible2012, donatiDeepGeometricFunctional2020} uses features $F_1$ and $F_2$ as \textit{functions} on the shapes, expected to correspond, and forces the functional map to effectively transfer them when solving the linear system in \Cref{eq:C linear}. On the other hand, the pointwise-map based branch solely relies on distances between rows of the feature matrices (\cref{eq:Kernel map}), viewing features as pointwise embeddings only.

Using a consistency loss between the two branches enables to merge the two effects, and, as highlighted in~\cite{sunSpatiallySpectrallyConsistent2023}, removing the spectral branch has a serious impact on the results. In our experiments in \Cref{sec:Results}, we observe that features learned without the spectral branch usually exhibited undesirable high-frequency variations. As emphasized by~\cite{attaikiUnderstandingImprovingFeatures2023}, smoothness of features is a key aspect for the generalization for functional map based methods. While all networks using the spectral branch provide relatively smooth features, we show in \Cref{subsec:res:learned features} that replacing this branch using a refinement consistency loss also promotes smoothness in features in our pipeline.

\section{Results}
\label{sec:Results}

In this section, we conduct a series of experiments to assess various aspects of our proposed method. In order to validate the capability of our entire pipeline, we first compare our method to several other works on multiple shape matching benchmarks.
We additionally wish to highlight our scalable dense maps appear as a valuable tool for many functional map based networks using dense pointwise maps, independently of our complete pipeline.
We therefore emphasize the memory scalability of our GPU-based ZoomOut algorithm compared to existing implementations of the algorithm.

Finally, inspired by~\cite{attaikiUnderstandingImprovingFeatures2023} we analyze how our novel ZoomOut consistency loss we introduced at train-time influences the features learned by our feature extractor.

\subsection{Datasets}

We evaluate the shape matching performance of our algorithm across four widely-used human datasets, commonly employed as benchmarks.
The evaluation includes the remeshed~\cite{renContinuousOrientationpreservingCorrespondences2019} version FAUST dataset~\cite{bogoFAUSTDatasetEvaluation2014} which contains 100 shapes, split in 80 and 20 shapes for training and testing as introduced in~\cite{donatiDeepGeometricFunctional2020}.
We also use the remeshed~\cite{renContinuousOrientationpreservingCorrespondences2019} version of the SCAPE dataset~\cite{anguelovSCAPEShapeCompletion2005} with 71 humans divided in 51 shapes for training and 20 for testing. For testing purposes only, the remeshed version of the SHREC19 dataset~\cite{melziMatchingHumansDifferent2019}, composed of 44 shapes, is also included.

While these datasets mostly contain near-isometric shapes, we also evaluate our method on the remeshed~\cite{magnetSmoothNonRigidShape2022} Deforming Things 4D dataset~\cite{li4DCompleteNonRigidMotion2021}, a challenging non-isometric dataset of humanoid shapes. In particular, we focus on the adapted version DT4D-H defined in~\cite{liLearningMultiresolutionFunctional2022}, which defines 198 shapes for training and 95 for testing.
Results on the SMAL~\cite{zuffi3DMenagerieModeling2017} dataset, with PCK curves, can be found in the supplementary material.

\subsection{Shape Matching Results}

In this work, we exclusively evaluate unsupervised learning performances, and therefore discard baselines focusing on pure supervised learning~\cite{trappoliniShapeRegistrationTime2021, litanyDeepFunctionalMaps2017, groueix3DCODED3DCorrespondences2018, wiersmaCNNsSurfacesUsing2020}.
As a reference, we provide results using axiomatic functional map algorithms such as ZoomOut~\cite{melziZoomOutSpectralUpsampling2019}, Discrete Optimization~\cite{renDiscreteOptimizationShape2021}, BCICP~\cite{renContinuousOrientationpreservingCorrespondences2019} and SmoothShells~\cite{eisenbergerSmoothShellsMultiScale2020}.

Our method can directly be compared to the following baselines: GeomFmaps~\cite{donatiDeepGeometricFunctional2020}, 
DUO-FMNet~\cite{donatiDeepOrientationawareFunctional2022}, DeepShells~\cite{eisenbergerDeepShellsUnsupervised2020}, NeuroMorph~\cite{eisenbergerNeuroMorphUnsupervisedShape}, AttentiveFMaps~\cite{liLearningMultiresolutionFunctional2022}, ConsistentFMaps~\cite{sunSpatiallySpectrallyConsistent2023}, UDMSM~\cite{caoUnsupervisedDeepMultishape2022}, and ULRSSM~\cite{caoUnsupervisedLearningRobust2023}.
Note that we all results are presented without test time refinement for fairness. In particular, ULRSSM~\cite{caoUnsupervisedLearningRobust2023} relies on fine-tuning the network for each shape in the test dataset independently, which we turn off to obtain the result. We provide results with fine-tuning using the ``w/ fine-tune'' tag.  Note that we provide comparison with a more complete set of methods in the supplementary materials, as well as results of our pipeline without using the consistency loss.

\Cref{tab:res:all humans} provides the mean geodesic error for all the aforementioned baselines, as well as for our pipeline described in \Cref{subsec:Pipeline}. We evaluate our methods on combinations of the Faust (\textbf{F}), Scape (\textbf{S}) and Shrec19 (\textbf{S19}), when training either on Faust and Scape independently, or jointly (\textbf{F+S}). This table shows our simple pipeline provides similar performance to state of the arts methods, all the while being greatly scalable to large meshes and removing the need for differentiation through a linear system solver.

In addition, we evaluate our network on the DeformingThings4D dataset, and in particular on the subset provided in~\cite{liLearningMultiresolutionFunctional2022} for evaluation, as displayed on \Cref{tab:res:dt4d}. Our method achieves better performance than existing 
baselines  on the inter-class category, which shows its capabilities even in non-isometric scenarios.

\begin{table}
\centering
{\def\arraystretch{1}\tabcolsep=0.5em

\begin{tabular}{ccc}\toprule[0.8pt]
Train & \multicolumn{2}{c}{\textbf{DT4D-H}} \\ \cmidrule(lr){2-3} 
Test & intra-class & inter-class \\ \midrule[0.8pt]
Deep Shells~\cite{eisenbergerDeepShellsUnsupervised2020} & 3.4 & 31.1\\
DUO-FMNet~\cite{donatiDeepOrientationawareFunctional2022} & 2.6  &15.8 \\
AttentiveFMaps~\cite{liLearningMultiresolutionFunctional2022} & 1.2 & 14.6 \\
ULRSSM~\cite{caoUnsupervisedLearningRobust2023} & \textbf{0.9} & 4.4   \\
ConsistentFMaps~\cite{sunSpatiallySpectrallyConsistent2023} & 1.2 & 6.1 \\ \midrule[0.5pt]
Ours & 1.8 &  \textbf{4.1}\\
\bottomrule[0.8pt]
\end{tabular}}
\caption{Mean geodesic error ($\times 100$) on the DeformingThing4D dataset subset from~\cite{liLearningMultiresolutionFunctional2022} (\textbf{DT4D-H}). Best results are highlighted in bold.} \label{tab:res:dt4d}\vspace{-12pt}
\end{table}

\begin{figure}[!h]
  \centering
  \includegraphics[width=0.8\linewidth]{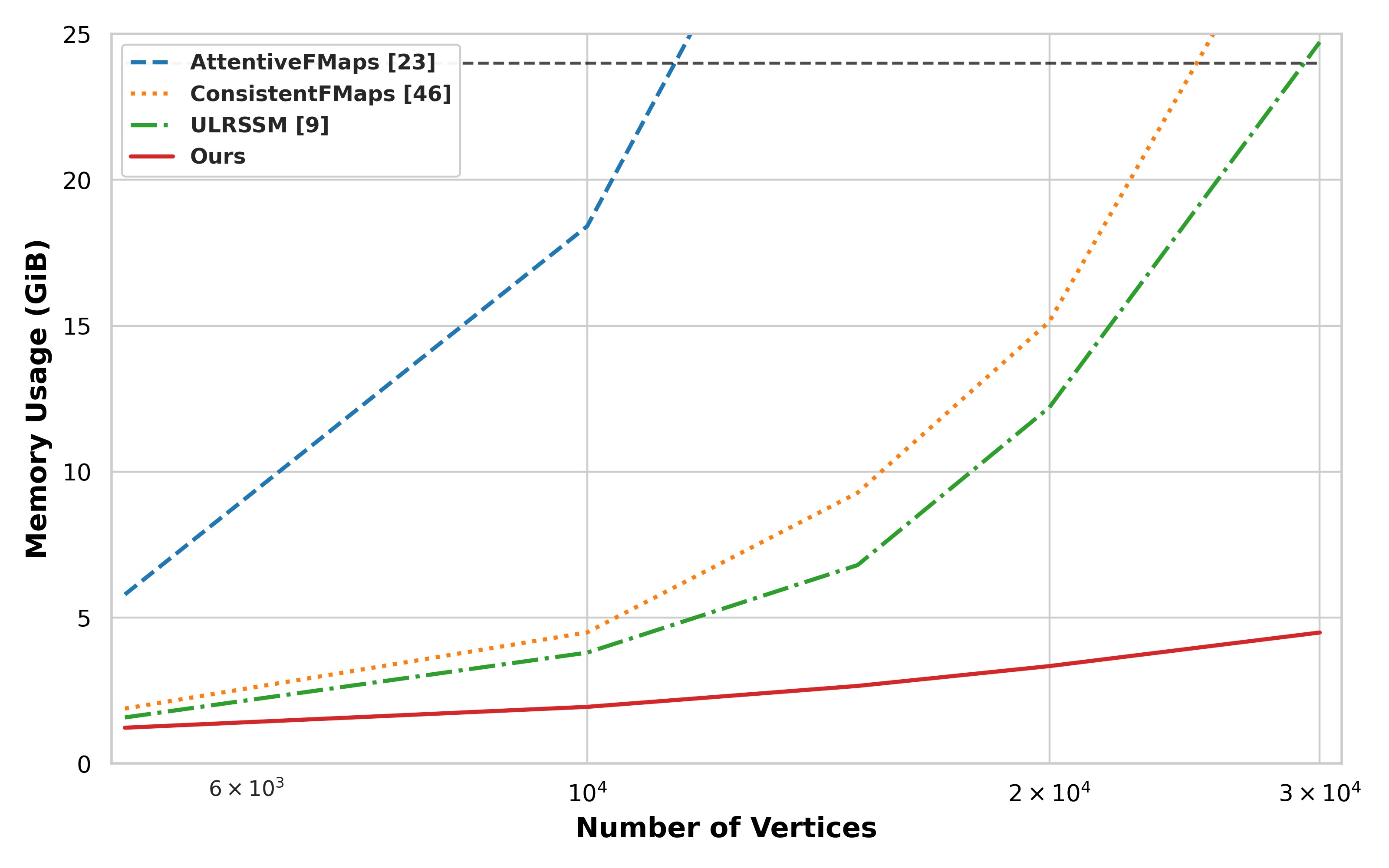}
   \caption{GPU memory usage when processing a single pair of shapes, depending on their vertex count.  Note, e.g., that AttentiveFMaps~\cite{liLearningMultiresolutionFunctional2022} runs out of 24GB memory after $11$k vertices.}
   \label{fig:memory benchmark} \vspace{-12pt}
\end{figure}
\subsection{Scalability to Dense Meshes}

In this section, we discuss the memory efficiency of our scalable maps, and highlight its speed performance in the case of very dense meshes where standard methods would go out of GPU-memory.

A first observation, provided in \Cref{fig:memory benchmark} shows the GPU memory usage, using varying number of vertices, of current state-of-the-art methods for unsupervised shape matching. In particular, we notice that AttentiveFMaps~\cite{liLearningMultiresolutionFunctional2022}, due to its multiple dense pointwise map computations, quickly runs out of 24 GiB GPU memory. On the other hand, while our method uses 11 different pointwise maps, its memory footprint remains significantly lower than competing methods~\cite{sunSpatiallySpectrallyConsistent2023, caoUnsupervisedLearningRobust2023}, in particular for large number of vertices.

Secondly, we analyze our results on the standard axiomatic ZoomOut algorithm~\cite{melziZoomOutSpectralUpsampling2019}, often used independently of learning pipelines, \eg as a means to obtain maps from simple landmarks. In that case, we observe that usual implementations never leverage GPU acceleration and were only run on CPU, and we easily ported the code to GPU using PyTorch.

In \Cref{tab:res:speedtest}, we first compare the CPU, GPU, and our version of ZoomOut, and show that the processing time in the presence of dense meshes remain reasonable.  Our version of ZoomOut (``Our ZoomOut'') uses the same tools used to implement our Differentiable ZoomOut in~\Cref{subsec:Differentiable ZoomOut}, with a scalable version of brute force nearest neighbor in Keops~\cite{Keops2021}, which again does not require fitting the distance matrix in memory.
We additionally compare a naïve PyTorch implementation of our Differentiable ZoomOut (\cref{subsec:Differentiable ZoomOut}) with one using our scalable dense maps. Finally, we add results by porting the approximation from~\cite{magnetScalableEfficientFunctional2023} to GPU and using scalable dense maps 
 (``Our + \cite{magnetScalableEfficientFunctional2023}''). More details on this mix and its usage are provided in the supplementary material.

\Cref{tab:res:speedtest} presents the results of applying these algorithms to shapes of varying sizes.  In the initial experiment with meshes of around 5000 vertices, all methods exhibit similar performance, significantly outperforming the CPU-based algorithm due to GPU utilization. However, with denser meshes containing $10^5$ vertices, conventional methods encounter GPU memory limitations, while our scalable dense maps offer notable improvements over existing approaches. Moreover, our modification  of~\cite{magnetScalableEfficientFunctional2023}, which approximates the algorithm, presents the fastest results without memory overloading. This solves the main speed bottleneck presented in~\cite{magnetScalableEfficientFunctional2023}, with more details provided in the supplementary.

Our method therefore allows using several dense pointwise maps simultaneously, or training and testing functional maps network directly on dense shapes. We refer the interested reader to the supplementary material for such experiments on dense meshes, including texture transfer visualization.

\begin{table}
\centering
{\def\arraystretch{1}\tabcolsep=0.5em
\begin{tabular}{ccc}\toprule[0.8pt]

 & Sparse (5K) & Dense (100k) \\ \midrule[0.8pt]
CPU ZoomOut & 3.6 s & 700 s\\
GPU ZoomOut & 0.1 s & OOM \\
GPU Diff. ZoomOut & 0.1 s & OOM \\
Our ZoomOut & 0.1 s  & 2.4 s \\
Our Diff. ZoomOut & 0.1 s & 5 s \\
Our + \cite{magnetScalableEfficientFunctional2023} & 0.1 s & 0.4 s\\
\bottomrule[0.8pt]
\end{tabular}}
\caption{Average processing time in seconds, between CPU, GPU, and memory scalable implementations of ZoomOut and Differentiable ZoomOut. The number of vertices is given in parentheses.} \label{tab:res:speedtest} \vspace{-12pt}
\end{table}

\subsection{Learned Features}
\label{subsec:res:learned features}


As highlighted in~\cite{attaikiUnderstandingImprovingFeatures2023}, analyzing the features learned in deep functional map networks  valuable insights into their performance. In particular, it was shown that achieving smooth features positively impacts the network's generalization capabilities.

The authors of~\cite{attaikiUnderstandingImprovingFeatures2023} thus advocated explicitly enforcing features smoothness using spectral projection. This was used in~\cite{sunSpatiallySpectrallyConsistent2023} as well as in AttentiveFMaps~\cite{liLearningMultiresolutionFunctional2022}. In contrast, we do not enforce such constraints and no no loss in our pipeline directly promotes smoothness. In particular, the dense pointwise map $\Pi$ built from the features do not use any neighboring information.

\begin{figure}
    \centering
    \includegraphics[width=.9\linewidth]{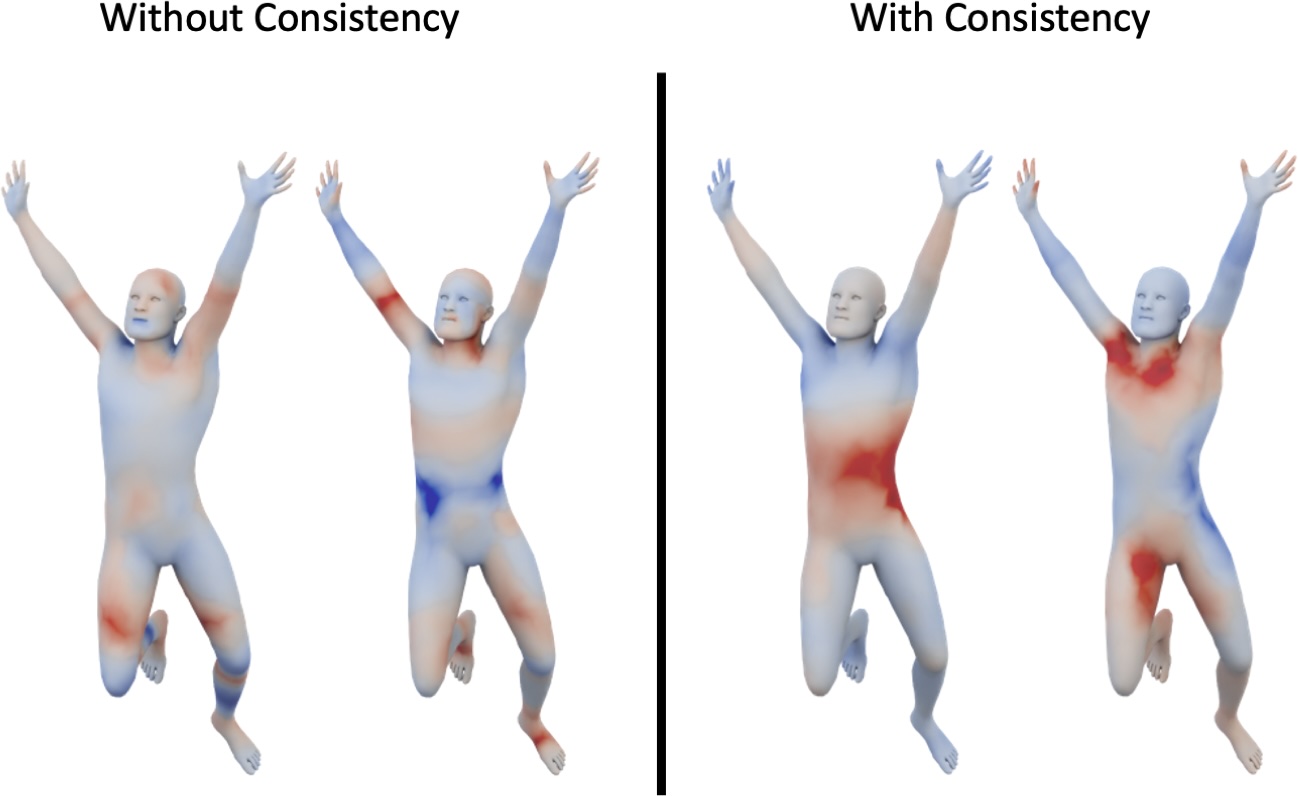}
    \caption{Example of feature functions learned by our model, with or without consistency loss. As noted by~\cite{attaikiUnderstandingImprovingFeatures2023}, smoother features are generally preferred for generalization purposes.}
    \label{fig:features} \vspace{-12pt}
\end{figure}

However, we show that the consistency loss introduced in \Cref{subsec:Differentiable ZoomOut} actually pushes the feature extractor to learn smooth features. To observe this, we retrain our network on the Scape dataset while removing the consistency loss from \Cref{eq:loss consist}, and visualize the learned features on test datasets. \Cref{fig:features} shows example of feature functions produced by the networks when trained with and without the consistency loss on a random surface from the SHREC19 dataset.
On the left side of this image, we observe that without refinement consistency, the features seem to highlight multiple small patches on the surface. In contrast, the feature functions learned by our method, displayed on the rightmost part of the image, present nicer patterns where large geodesic patches of the surfaces are highlighted.

We argue that obtaining an orthogonal functional map from a soft pointwise map built with features does not require such features to exhibit smoothness. However, the introduction of the consistency loss serves a dual purpose. While its primary role is to align the functional map with the output of a refinement algorithm, it inadvertently acts as a compelling constraint that encourages the learning of smoother features. As this property has been noted as key to performance~\cite{attaikiUnderstandingImprovingFeatures2023, sunSpatiallySpectrallyConsistent2023}, these result leads us to believe that incorporating such a loss into existing pipelines holds significant promise for enhancing overall performance in functional map learning.

\section{Conclusion, Limitations \& Future Work}
In this work, we presented a novel approach to compute functional maps using soft pointwise map, without ever storing the dense matrix in memory. This novel implementation enables use to derive a fast, differentiable and memory efficient version of the ZoomOut algorithm~\cite{melziZoomOutSpectralUpsampling2019}. In turn, we use this algorithm while training and derive a new consistency loss between the initial and refined version of the predicted functional map. We notice this loss appears particularly effective and allows us to use a new single-branch architecture for functional map learning, which does not require differentiating through a linear system.

One major limitation of our method is its dependence on the computation of the spectrum Laplacian of the input shapes, which can become prohibitively slow with larger shapes. Furthermore, the ZoomOut algorithm, while particularly fit to handle near-isometric shapes, is prone to fail in the presence of highly non-isometric deformations or partiality~\cite{magnetSmoothNonRigidShape2022}. The guidance provided by the consistency loss would then be unfit for the problem.

Future research could therefore seek to handle meshes with higher differences such as partiality, noise or simply high distortion. This would potentially require incorporating other refinement algorithms into the training pipeline. Investigating the impact of our new consistency loss in various pipelines would also contribute to a comprehensive understanding of its applicability and effectiveness.

\paragraph*{Acknowledgements.} The authors thank the anonymous reviewers for their valuable comments and suggestions. Parts of this work were supported by the ERC Starting Grant 758800 (EXPROTEA), ERC Consolidator Grant 101087347 (VEGA), ANR AI Chair AIGRETTE, as well as gifts from Ansys and Adobe Research.
{
    \small
    \bibliographystyle{ieeenat_fullname}
    \bibliography{cvpr23_biblio}

\begin{thebibliography}{52}
\providecommand{\natexlab}[1]{#1}
\providecommand{\url}[1]{\texttt{#1}}
\expandafter\ifx\csname urlstyle\endcsname\relax
  \providecommand{\doi}[1]{doi: #1}\else
  \providecommand{\doi}{doi: \begingroup \urlstyle{rm}\Url}\fi

\bibitem[Anguelov et~al.(2005)Anguelov, Srinivasan, Koller, Thrun, Rodgers, and Davis]{anguelovSCAPEShapeCompletion2005}
Dragomir Anguelov, Praveen Srinivasan, Daphne Koller, Sebastian Thrun, Jim Rodgers, and James Davis.
\newblock {{SCAPE}}: Shape completion and animation of people.
\newblock \emph{ACM Transactions on Graphics}, 24\penalty0 (3):\penalty0 408--416, 2005.

\bibitem[Attaiki and Ovsjanikov(2023)]{attaikiUnderstandingImprovingFeatures2023}
Souhaib Attaiki and Maks Ovsjanikov.
\newblock Understanding and {{Improving Features Learned}} in {{Deep Functional Maps}}.
\newblock In \emph{2023 {{IEEE}}/{{CVF Conference}} on {{Computer Vision}} and {{Pattern Recognition}} ({{CVPR}})}, pages 1316--1326, {Vancouver, BC, Canada}, 2023. {IEEE}.

\bibitem[Attaiki et~al.(2021)Attaiki, Pai, and Ovsjanikov]{attaikiDPFMDeepPartial2021}
Souhaib Attaiki, Gautam Pai, and Maks Ovsjanikov.
\newblock {{DPFM}}: {{Deep Partial Functional Maps}}.
\newblock In \emph{2021 {{International Conference}} on {{3D Vision}} ({{3DV}})}, pages 175--185, {London, United Kingdom}, 2021. {IEEE}.

\bibitem[Aubry et~al.(2011)Aubry, Schlickewei, and Cremers]{aubryWaveKernelSignature2011}
Mathieu Aubry, Ulrich Schlickewei, and Daniel Cremers.
\newblock The wave kernel signature: {{A}} quantum mechanical approach to shape analysis.
\newblock In \emph{2011 {{IEEE International Conference}} on {{Computer Vision Workshops}} ({{ICCV Workshops}})}, pages 1626--1633, {Barcelona, Spain}, 2011. {IEEE}.

\bibitem[Bogo et~al.(2014)Bogo, Romero, Loper, and Black]{bogoFAUSTDatasetEvaluation2014}
Federica Bogo, Javier Romero, Matthew Loper, and Michael~J. Black.
\newblock {{FAUST}}: {{Dataset}} and {{Evaluation}} for {{3D Mesh Registration}}.
\newblock In \emph{Proceedings of the {{IEEE Conference}} on {{Computer Vision}} and {{Pattern Recognition}}}, pages 3794--3801, 2014.

\bibitem[Bogo et~al.(2017)Bogo, Romero, {Pons-Moll}, and Black]{bogoDynamicFAUSTRegistering2017}
Federica Bogo, Javier Romero, Gerard {Pons-Moll}, and Michael~J. Black.
\newblock Dynamic {{FAUST}}: {{Registering Human Bodies}} in {{Motion}}.
\newblock In \emph{2017 {{IEEE Conference}} on {{Computer Vision}} and {{Pattern Recognition}} ({{CVPR}})}, pages 5573--5582, {Honolulu, HI}, 2017. {IEEE}.

\bibitem[Bronstein and Kokkinos(2010)]{bronsteinScaleinvariantHeatKernel2010}
Michael~M. Bronstein and Iasonas Kokkinos.
\newblock Scale-invariant heat kernel signatures for non-rigid shape recognition.
\newblock In \emph{2010 {{IEEE Computer Society Conference}} on {{Computer Vision}} and {{Pattern Recognition}}}, pages 1704--1711, {San Francisco, CA, USA}, 2010. {IEEE}.

\bibitem[Cao and Bernard(2022)]{caoUnsupervisedDeepMultishape2022}
Dongliang Cao and Florian Bernard.
\newblock Unsupervised {{Deep Multi-shape Matching}}.
\newblock In \emph{Computer {{Vision}} \textendash{} {{ECCV}} 2022}, pages 55--71. {Springer Nature Switzerland}, {Cham}, 2022.

\bibitem[Cao et~al.(2023)Cao, Roetzer, and Bernard]{caoUnsupervisedLearningRobust2023}
Dongliang Cao, Paul Roetzer, and Florian Bernard.
\newblock Unsupervised {{Learning}} of {{Robust Spectral Shape Matching}}.
\newblock \emph{ACM Transactions on Graphics}, 42\penalty0 (4):\penalty0 132:1--132:15, 2023.

\bibitem[Charlier et~al.(2021)Charlier, Feydy, Glaunès, Collin, and Durif]{Keops2021}
Benjamin Charlier, Jean Feydy, Joan~Alexis Glaunès, François-David Collin, and Ghislain Durif.
\newblock Kernel operations on the gpu, with autodiff, without memory overflows.
\newblock \emph{Journal of Machine Learning Research}, 22\penalty0 (74):\penalty0 1--6, 2021.

\bibitem[Deng et~al.(2022)Deng, Yao, Dyke, and Zhang]{dengSurveyNonRigid3D2022}
Bailin Deng, Yuxin Yao, Roberto~M. Dyke, and Juyong Zhang.
\newblock A {{Survey}} of {{Non-Rigid 3D Registration}}.
\newblock \emph{Computer Graphics Forum}, 41\penalty0 (2):\penalty0 559--589, 2022.

\bibitem[Donati et~al.(2020)Donati, Sharma, and Ovsjanikov]{donatiDeepGeometricFunctional2020}
Nicolas Donati, Abhishek Sharma, and Maks Ovsjanikov.
\newblock Deep {{Geometric Functional Maps}}: {{Robust Feature Learning}} for {{Shape Correspondence}}.
\newblock In \emph{2020 {{IEEE}}/{{CVF Conference}} on {{Computer Vision}} and {{Pattern Recognition}} ({{CVPR}})}, pages 8589--8598, {Seattle, WA, USA}, 2020. {IEEE}.

\bibitem[Donati et~al.(2022{\natexlab{a}})Donati, Corman, Melzi, and Ovsjanikov]{donatiComplexFunctionalMaps2022a}
Nicolas Donati, Etienne Corman, Simone Melzi, and Maks Ovsjanikov.
\newblock Complex {{Functional Maps}}: {{A Conformal Link Between Tangent Bundles}}.
\newblock \emph{Computer Graphics Forum}, 41\penalty0 (1):\penalty0 317--334, 2022{\natexlab{a}}.

\bibitem[Donati et~al.(2022{\natexlab{b}})Donati, Corman, and Ovsjanikov]{donatiDeepOrientationawareFunctional2022}
Nicolas Donati, Etienne Corman, and Maks Ovsjanikov.
\newblock Deep orientation-aware functional maps: {{Tackling}} symmetry issues in {{Shape Matching}}.
\newblock In \emph{2022 {{IEEE}}/{{CVF Conference}} on {{Computer Vision}} and {{Pattern Recognition}} ({{CVPR}})}, pages 732--741, {New Orleans, LA, USA}, 2022{\natexlab{b}}. {IEEE}.

\bibitem[Efroni et~al.(2022)Efroni, Ginzburg, and Raviv]{efroniSpectralTeacherSpatial2022}
Omri Efroni, Dvir Ginzburg, and Dan Raviv.
\newblock Spectral {{Teacher}} for a {{Spatial Student}}: {{Spectrum-Aware Real-Time Dense Shape Correspondence}}.
\newblock In \emph{2022 {{International Conference}} on {{3D Vision}} ({{3DV}})}, pages 1--10, 2022.

\bibitem[Eisenberger et~al.(2020{\natexlab{a}})Eisenberger, Lahner, and Cremers]{eisenbergerSmoothShellsMultiScale2020}
Marvin Eisenberger, Zorah Lahner, and Daniel Cremers.
\newblock Smooth {{Shells}}: {{Multi-Scale Shape Registration With Functional Maps}}.
\newblock In \emph{Proceedings of the {{IEEE}}/{{CVF Conference}} on {{Computer Vision}} and {{Pattern Recognition}}}, pages 12265--12274, 2020{\natexlab{a}}.

\bibitem[Eisenberger et~al.(2020{\natexlab{b}})Eisenberger, Toker, {Leal-Taix{\'e}}, and Cremers]{eisenbergerDeepShellsUnsupervised2020}
Marvin Eisenberger, Aysim Toker, Laura {Leal-Taix{\'e}}, and Daniel Cremers.
\newblock Deep {{Shells}}: {{Unsupervised Shape Correspondence}} with {{Optimal Transport}}.
\newblock \emph{Advances in Neural Information Processing Systems}, 33, 2020{\natexlab{b}}.

\bibitem[Eisenberger et~al.(2021)Eisenberger, Novotny, Kerchenbaum, Labatut, Neverova, Cremers, and Vedaldi]{eisenbergerNeuroMorphUnsupervisedShape}
M. Eisenberger, D. Novotny, G. Kerchenbaum, P. Labatut, N. Neverova, D. Cremers, and A. Vedaldi.
\newblock {{NeuroMorph}}: {{Unsupervised Shape Interpolation}} and {{Correspondence}} in {{One Go}}.
\newblock In \emph{2021 IEEE/CVF Conference on Computer Vision and Pattern Recognition (CVPR)}, pages 7469--7479, Los Alamitos, CA, USA, 2021. IEEE Computer Society.

\bibitem[Ezuz and {Ben-Chen}(2017)]{ezuzDeblurringDenoisingMaps2017}
Danielle Ezuz and Mirela {Ben-Chen}.
\newblock Deblurring and {{Denoising}} of {{Maps}} between {{Shapes}}.
\newblock \emph{Computer Graphics Forum}, 36\penalty0 (5):\penalty0 165--174, 2017.

\bibitem[Groueix et~al.(2018)Groueix, Fisher, Kim, Russell, and Aubry]{groueix3DCODED3DCorrespondences2018}
Thibault Groueix, Matthew Fisher, Vladimir~G. Kim, Bryan~C. Russell, and Mathieu Aubry.
\newblock {{3D-CODED}} : {{3D Correspondences}} by {{Deep Deformation}}.
\newblock \emph{arXiv:1806.05228 [cs]}, 2018.

\bibitem[Johnson et~al.(2019)Johnson, Douze, and J{\'e}gou]{johnson2019billion}
Jeff Johnson, Matthijs Douze, and Herv{\'e} J{\'e}gou.
\newblock Billion-scale similarity search with {GPUs}.
\newblock \emph{IEEE Transactions on Big Data}, 7\penalty0 (3):\penalty0 535--547, 2019.

\bibitem[Kingma and Ba(2015)]{kingmaAdamOptimizer2015}
Diederik Kingma and Jimmy Ba.
\newblock Adam: A method for stochastic optimization.
\newblock In \emph{International Conference on Learning Representations (ICLR)}, San Diega, CA, USA, 2015.

\bibitem[Li et~al.(2022)Li, Donati, and Ovsjanikov]{liLearningMultiresolutionFunctional2022}
Lei Li, Nicolas Donati, and Maks Ovsjanikov.
\newblock Learning {{Multi-resolution Functional Maps}} with {{Spectral Attention}} for {{Robust Shape Matching}}.
\newblock \emph{Advances in Neural Information Processing Systems}, 35:\penalty0 29336--29349, 2022.

\bibitem[Li et~al.(2021)Li, Takehara, Taketomi, Zheng, and Niesner]{li4DCompleteNonRigidMotion2021}
Yang Li, Hikari Takehara, Takafumi Taketomi, Bo Zheng, and Matthias Niesner.
\newblock {{4DComplete}}: {{Non-Rigid Motion Estimation Beyond}} the {{Observable Surface}}.
\newblock In \emph{2021 {{IEEE}}/{{CVF International Conference}} on {{Computer Vision}} ({{ICCV}})}, pages 12686--12696, {Montreal, QC, Canada}, 2021. {IEEE}.

\bibitem[Litany et~al.(2017)Litany, Remez, Rodola, Bronstein, and Bronstein]{litanyDeepFunctionalMaps2017}
Or Litany, Tal Remez, Emanuele Rodola, Alex Bronstein, and Michael Bronstein.
\newblock Deep {{Functional Maps}}: {{Structured Prediction}} for {{Dense Shape Correspondence}}.
\newblock In \emph{2017 {{IEEE International Conference}} on {{Computer Vision}} ({{ICCV}})}, pages 5660--5668, {Venice}, 2017. {IEEE}.

\bibitem[Magnet and Ovsjanikov(2023)]{magnetScalableEfficientFunctional2023}
Robin Magnet and Maks Ovsjanikov.
\newblock Scalable and {{Efficient Functional Map Computations}} on {{Dense Meshes}}.
\newblock \emph{Computer Graphics Forum}, 42\penalty0 (2):\penalty0 89--101, 2023.

\bibitem[Magnet et~al.(2022)Magnet, Ren, {Sorkine-Hornung}, and Ovsjanikov]{magnetSmoothNonRigidShape2022}
Robin Magnet, Jing Ren, Olga {Sorkine-Hornung}, and Maks Ovsjanikov.
\newblock Smooth {{Non-Rigid Shape Matching}} via {{Effective Dirichlet Energy Optimization}}.
\newblock In \emph{2022 {{International Conference}} on {{3D Vision}} ({{3DV}})}, pages 495--504, {Prague, Czech Republic}, 2022. {IEEE}.

\bibitem[Meanti et~al.(2020)Meanti, Carratino, Rosasco, and Rudi]{meantiKernelMethodsRoof2020}
Giacomo Meanti, Luigi Carratino, Lorenzo Rosasco, and Alessandro Rudi.
\newblock Kernel {{Methods Through}} the {{Roof}}: {{Handling Billions}} of {{Points Efficiently}}.
\newblock In \emph{Advances in {{Neural Information Processing Systems}}}, pages 14410--14422. {Curran Associates, Inc.}, 2020.

\bibitem[Melzi et~al.(2019{\natexlab{a}})Melzi, Marin, Rodol{\`a}, Castellani, Ren, Poulenard, Wonka, and Ovsjanikov]{melziMatchingHumansDifferent2019}
S. Melzi, R. Marin, E. Rodol{\`a}, U. Castellani, J. Ren, A. Poulenard, P. Wonka, and M. Ovsjanikov.
\newblock \emph{Matching {{Humans}} with {{Different Connectivity}}}.
\newblock {The Eurographics Association}, 2019{\natexlab{a}}.

\bibitem[Melzi et~al.(2019{\natexlab{b}})Melzi, Ren, Rodol{\`a}, Sharma, Wonka, and Ovsjanikov]{melziZoomOutSpectralUpsampling2019}
Simone Melzi, Jing Ren, Emanuele Rodol{\`a}, Abhishek Sharma, Peter Wonka, and Maks Ovsjanikov.
\newblock {{ZoomOut}}: Spectral upsampling for efficient shape correspondence.
\newblock \emph{ACM Transactions on Graphics}, 38\penalty0 (6):\penalty0 1--14, 2019{\natexlab{b}}.

\bibitem[Meyer et~al.(2003)Meyer, Desbrun, Schr{\"o}der, and Barr]{meyerDiscreteDifferentialGeometryOperators2003}
Mark Meyer, Mathieu Desbrun, Peter Schr{\"o}der, and Alan~H. Barr.
\newblock Discrete {{Differential-Geometry Operators}} for {{Triangulated}} 2-{{Manifolds}}.
\newblock In \emph{Visualization and {{Mathematics III}}}, pages 35--57. {Springer Berlin Heidelberg}, {Berlin, Heidelberg}, 2003.

\bibitem[Nogneng and Ovsjanikov(2017)]{nognengInformativeDescriptorPreservation2017}
Dorian Nogneng and Maks Ovsjanikov.
\newblock Informative {{Descriptor Preservation}} via {{Commutativity}} for {{Shape Matching}}.
\newblock \emph{Computer Graphics Forum}, 36\penalty0 (2):\penalty0 259--267, 2017.

\bibitem[Nogneng et~al.(2018)Nogneng, Melzi, Rodol{\`a}, Castellani, Bronstein, and Ovsjanikov]{nognengImprovedFunctionalMappings2018}
D. Nogneng, S. Melzi, E. Rodol{\`a}, U. Castellani, M. Bronstein, and M. Ovsjanikov.
\newblock Improved {{Functional Mappings}} via {{Product Preservation}}.
\newblock \emph{Computer Graphics Forum}, 37\penalty0 (2):\penalty0 179--190, 2018.

\bibitem[Ovsjanikov et~al.(2012)Ovsjanikov, {Ben-Chen}, Solomon, Butscher, and Guibas]{ovsjanikovFunctionalMapsFlexible2012}
Maks Ovsjanikov, Mirela {Ben-Chen}, Justin Solomon, Adrian Butscher, and Leonidas Guibas.
\newblock Functional maps: A flexible representation of maps between shapes.
\newblock \emph{ACM Transactions on Graphics}, 31\penalty0 (4):\penalty0 1--11, 2012.

\bibitem[Ovsjanikov et~al.(2017)Ovsjanikov, Corman, Bronstein, Rodol{\`a}, {Ben-Chen}, Guibas, Chazal, and Bronstein]{ovsjanikovComputingProcessingCorrespondences2017}
Maks Ovsjanikov, Etienne Corman, Michael Bronstein, Emanuele Rodol{\`a}, Mirela {Ben-Chen}, Leonidas Guibas, Frederic Chazal, and Alex Bronstein.
\newblock Computing and processing correspondences with functional maps.
\newblock In \emph{{{ACM SIGGRAPH}} 2017 {{Courses}}}, pages 1--62, {New York, NY, USA}, 2017. {Association for Computing Machinery}.

\bibitem[Pai et~al.(2021)Pai, Ren, Melzi, Wonka, and Ovsjanikov]{paiFastSinkhornFilters2021}
Gautam Pai, Jing Ren, Simone Melzi, Peter Wonka, and Maks Ovsjanikov.
\newblock Fast {{Sinkhorn Filters}}: {{Using Matrix Scaling}} for {{Non-Rigid Shape Correspondence}} with {{Functional Maps}}.
\newblock In \emph{{{CVPR}}}, {Nashville (virtual), United States}, 2021.

\bibitem[Ren et~al.(2019{\natexlab{a}})Ren, Panine, Wonka, and Ovsjanikov]{renStructuredRegularizationFunctional2019}
Jing Ren, Mikhail Panine, Peter Wonka, and Maks Ovsjanikov.
\newblock Structured {{Regularization}} of {{Functional Map Computations}}.
\newblock \emph{Computer Graphics Forum}, 38\penalty0 (5):\penalty0 39--53, 2019{\natexlab{a}}.

\bibitem[Ren et~al.(2019{\natexlab{b}})Ren, Poulenard, Wonka, and Ovsjanikov]{renContinuousOrientationpreservingCorrespondences2019}
Jing Ren, Adrien Poulenard, Peter Wonka, and Maks Ovsjanikov.
\newblock Continuous and orientation-preserving correspondences via functional maps.
\newblock \emph{ACM Transactions on Graphics}, 37\penalty0 (6):\penalty0 1--16, 2019{\natexlab{b}}.

\bibitem[Ren et~al.(2021)Ren, Melzi, Wonka, and Ovsjanikov]{renDiscreteOptimizationShape2021}
Jing Ren, Simone Melzi, Peter Wonka, and Maks Ovsjanikov.
\newblock Discrete {{Optimization}} for {{Shape Matching}}.
\newblock \emph{Computer Graphics Forum}, 40\penalty0 (5):\penalty0 81--96, 2021.

\bibitem[Roufosse et~al.(2019)Roufosse, Sharma, and Ovsjanikov]{roufosseUnsupervisedDeepLearning2019}
Jean-Michel Roufosse, Abhishek Sharma, and Maks Ovsjanikov.
\newblock Unsupervised {{Deep Learning}} for {{Structured Shape Matching}}.
\newblock In \emph{2019 {{IEEE}}/{{CVF International Conference}} on {{Computer Vision}} ({{ICCV}})}, pages 1617--1627, {Seoul, Korea (South)}, 2019. {IEEE}.

\bibitem[Rudi et~al.(2017)Rudi, Carratino, and Rosasco]{rudiFALKONOptimalLarge2017}
Alessandro Rudi, Luigi Carratino, and Lorenzo Rosasco.
\newblock {{FALKON}}: {{An Optimal Large Scale Kernel Method}}.
\newblock In \emph{Advances in {{Neural Information Processing Systems}}}. {Curran Associates, Inc.}, 2017.

\bibitem[Sahillio{\u g}lu(2020)]{sahilliogluRecentAdvancesShape2020}
Yusuf Sahillio{\u g}lu.
\newblock Recent advances in shape correspondence.
\newblock \emph{The Visual Computer}, 36\penalty0 (8):\penalty0 1705--1721, 2020.

\bibitem[Sharp et~al.(2022)Sharp, Attaiki, Crane, and Ovsjanikov]{sharpDiffusionNetDiscretizationAgnostic2022a}
Nicholas Sharp, Souhaib Attaiki, Keenan Crane, and Maks Ovsjanikov.
\newblock {{DiffusionNet}}: {{Discretization Agnostic Learning}} on {{Surfaces}}.
\newblock \emph{ACM Transactions on Graphics}, 41\penalty0 (3):\penalty0 27:1--27:16, 2022.

\bibitem[Shi et~al.(2020)Shi, Xu, Yuan, and Fang]{shiUnsupervisedDeepShape2020}
Yi Shi, Mengchen Xu, Shuaihang Yuan, and Yi Fang.
\newblock Unsupervised {{Deep Shape Descriptor With Point Distribution Learning}}.
\newblock In \emph{2020 {{IEEE}}/{{CVF Conference}} on {{Computer Vision}} and {{Pattern Recognition}} ({{CVPR}})}, pages 9350--9359, {Seattle, WA, USA}, 2020. {IEEE}.

\bibitem[Sorkine and Alexa(2007)]{sorkineAsRigidAsPossibleSurfaceModeling}
Olga Sorkine and Marc Alexa.
\newblock {As-Rigid-As-Possible Surface Modeling}.
\newblock In \emph{Geometry Processing}. The Eurographics Association, 2007.

\bibitem[Sun et~al.(2023)Sun, Mao, Jiang, Ovsjanikov, and Huang]{sunSpatiallySpectrallyConsistent2023}
Mingze Sun, Shiwei Mao, Puhua Jiang, Maks Ovsjanikov, and Ruqi Huang.
\newblock Spatially and {{Spectrally Consistent Deep Functional Maps}}.
\newblock In \emph{Proceedings of the {{IEEE}}/{{CVF International Conference}} on {{Computer Vision}}}, pages 14497--14507, 2023.

\bibitem[Thomas et~al.(2019)Thomas, Qi, Deschaud, Marcotegui, Goulette, and Guibas]{thomasKPConvFlexibleDeformable2019}
Hugues Thomas, Charles~R. Qi, Jean-Emmanuel Deschaud, Beatriz Marcotegui, Fran{\c c}ois Goulette, and Leonidas~J. Guibas.
\newblock {{KPConv}}: {{Flexible}} and {{Deformable Convolution}} for {{Point Clouds}}.
\newblock \emph{arXiv:1904.08889 [cs]}, 2019.

\bibitem[Tombari et~al.(2010)Tombari, Salti, and Di~Stefano]{tombariUniqueSignaturesHistograms2010}
Federico Tombari, Samuele Salti, and Luigi Di~Stefano.
\newblock Unique {{Signatures}} of {{Histograms}} for {{Local Surface Description}}.
\newblock In \emph{Computer {{Vision}} \textendash{} {{ECCV}} 2010}, pages 356--369, {Berlin, Heidelberg}, 2010. {Springer}.

\bibitem[Trappolini et~al.(2021)Trappolini, Cosmo, Moschella, Marin, Melzi, and Rodol{\`a}]{trappoliniShapeRegistrationTime2021}
Giovanni Trappolini, Luca Cosmo, Luca Moschella, Riccardo Marin, Simone Melzi, and Emanuele Rodol{\`a}.
\newblock Shape {{Registration}} in the {{Time}} of {{Transformers}}.
\newblock In \emph{Advances in {{Neural Information Processing Systems}}}, pages 5731--5744. {Curran Associates, Inc.}, 2021.

\bibitem[Vestner et~al.(2017)Vestner, Litman, Rodola, Bronstein, and Cremers]{vestnerProductManifoldFilter2017}
Matthias Vestner, Roee Litman, Emanuele Rodola, Alex Bronstein, and Daniel Cremers.
\newblock Product {{Manifold Filter}}: {{Non-rigid Shape Correspondence}} via {{Kernel Density Estimation}} in the {{Product Space}}.
\newblock In \emph{2017 {{IEEE Conference}} on {{Computer Vision}} and {{Pattern Recognition}} ({{CVPR}})}, pages 6681--6690, {Honolulu, HI}, 2017. {IEEE}.

\bibitem[Wiersma et~al.(2020)Wiersma, Eisemann, and Hildebrandt]{wiersmaCNNsSurfacesUsing2020}
Ruben Wiersma, Elmar Eisemann, and Klaus Hildebrandt.
\newblock {{CNNs}} on surfaces using rotation-equivariant features.
\newblock \emph{ACM Transactions on Graphics}, 39\penalty0 (4):\penalty0 92:92:1--92:92:12, 2020.

\bibitem[Zuffi et~al.(2017)Zuffi, Kanazawa, Jacobs, and Black]{zuffi3DMenagerieModeling2017}
Silvia Zuffi, Angjoo Kanazawa, David~W. Jacobs, and Michael~J. Black.
\newblock {{3D Menagerie}}: {{Modeling}} the {{3D Shape}} and {{Pose}} of {{Animals}}.
\newblock \emph{2017 IEEE Conference on Computer Vision and Pattern Recognition (CVPR)}, pages 5524--5532, 2017.

\end{thebibliography}
}

\clearpage
\maketitlesupplementary

\begin{table*}
\centering
{\def\arraystretch{1}\tabcolsep=1em

\begin{tabular}{cccccccccc}\toprule[0.8pt]
Train & \multicolumn{3}{c}{\textbf{F}} & \multicolumn{3}{c}{\textbf{S}} & \multicolumn{3}{c}{\textbf{F+S}} \\ \cmidrule(lr){2-4} \cmidrule(lr){5-7} \cmidrule(lr){8-10}
Test & F & S & S19 & F & S & S19 & F & S & S19  \\ \midrule[0.8pt]

BCICP~\cite{renContinuousOrientationpreservingCorrespondences2019} & 6.1 &- &- & - &11.& -& - &- &-\\
ZoomOut~\cite{melziZoomOutSpectralUpsampling2019} & 6.1 &- &- & - &7.5& - &-& -& -\\
SmoothShells~\cite{eisenbergerSmoothShellsMultiScale2020} & 2.5 &- &- &- &4.7 &- &-& -& -\\ 
DiscreteOp~\cite{renDiscreteOptimizationShape2021} &5.6 & - &- & - & 13.1 &- & - & - & -\\\midrule[0.8pt]
GeomFmaps~\cite{donatiDeepGeometricFunctional2020} & 3.5 & 4.8 & 8.5 & 4.0 & 4.3 & 11.2 & 3.5 & 4.4  & 7.1 \\
Deep Shells~\cite{eisenbergerDeepShellsUnsupervised2020} & 1.7 & 5.4 & 27.4 & 2.7 & 2.5 & 23.4 & 1.6 &  2.4 & 21.1 \\
NeuroMorph~\cite{eisenbergerNeuroMorphUnsupervisedShape} & 8.5 & 28.5 & 26.3 & 18.2 & 29.9 & 27.6 &  9.1 & 27.3 & 25.3 \\
DUO-FMNet~\cite{donatiDeepOrientationawareFunctional2022} & 2.5 &  4.2 & 6.4 & 2.7 & 2.6 & 8.4 & 2.5 & 4.3 & 6.4 \\
UDMSM~\cite{caoUnsupervisedDeepMultishape2022} & \textbf{1.5} & 7.3 & 21.5 & 8.6 & 2.0 & 30.7 & 1.7 & 3.2 & 17.8 \\
ULRSSM~\cite{caoUnsupervisedLearningRobust2023} &  1.6 & 6.4 & 14.5 & 4.5 & \textbf{1.8} & 18.5 & \textbf{1.5} & \textbf{2.0} & 7.9  \\
ULRSSM (w/ fine-tune)~\cite{caoUnsupervisedLearningRobust2023} & 1.6 & 2.2 & 5.7 & 1.6 & 1.9 & 6.7 & 1.6 & 2.1 & 4.6  \\
AttentiveFMaps Fast~\cite{liLearningMultiresolutionFunctional2022} & 1.9 & 2.6 & 5.8 &  1.9 & 2.1 & 8.1 & 1.9 & 2.3 & 6.3 \\
AttentiveFMaps~\cite{liLearningMultiresolutionFunctional2022} & 1.9 & 2.6 &  6.4 & 2.2 & 2.2 & 9.9 & 1.9 & 2.3 & 5.8 \\
ConsistentFMaps~\cite{sunSpatiallySpectrallyConsistent2023} & 2.3 & 2.6 & \textbf{3.8} &2.4 & 2.5 &\textbf{4.5} & 2.2 & 2.3 & 4.3\\
ConsistentFMaps (dim 80)~\cite{sunSpatiallySpectrallyConsistent2023} & 1.7 & 2.6 & 5.5 & 2.2 & 2.0 & 5.8 & 1.7 & 2.2 &  5.6 \\ \midrule[0,5pt]
Ours & 1.9 &\textbf{ 2.4} & 4.2 & \textbf{1.9} & 2.4 & 6.9 & 1.9 & 2.3 & \textbf{3.6}\\
\bottomrule[0.8pt]
\end{tabular}}
\caption{Mean geodesic errors ($\times 100$) when training and testing on the Faust, Scape and Shrec19 datasets. Due to the fine-tuning strategy on ULRSSM (w/ fine-tune), we do not hightlight its results. See text for details.}  \label{tab:suppl:baselines}
\end{table*}
\section{Implementation Details}
\label{suppl:sec:Implementation}
In this section we provide more detailed information on the implementation of our model described in \Cref{fig:pipeline,subsec:Pipeline}.

Our model takes as input shapes using $128$ WKS descriptors computed from $128$ eigenfunctions of the Laplace-Betlrami operator. Similarly to~\cite{liLearningMultiresolutionFunctional2022,caoUnsupervisedLearningRobust2023}, each descriptor function is normalized on the shape with respect to the standard $L^2$ inner product on a mesh.
These descriptors are then fed to a DiffusionBlock with 4 Diffusion blocks of width of $256$, in a standard manner~\cite{sharpDiffusionNetDiscretizationAgnostic2022a, attaikiUnderstandingImprovingFeatures2023, caoUnsupervisedLearningRobust2023, sunSpatiallySpectrallyConsistent2023}. The main difference with these implementations is that we only output $32$ feature functions instead of the $128$ or $256$ usually used.

Features produces by DiffusionNet are used in our Differentiable ZoomOut block which first normalizes the pointwise features, and then computes a scalable dense map equivalent to standard two-branch networks, as shown in~\Cref{sec:Notations and background,eq:Kernel map}. Using this map, an initial functional map $\*C_{\text{init}}$ of size $K_{\text{init}}=30$ is computed, and is fed into a ZoomOut algorithm~\cite{melziZoomOutSpectralUpsampling2019} for 10 iterations with a spectral upsampling step of 10, where the pointwise maps are replaced by our scalable dense maps. This eventually produces our refined map $\*C_{\text{refined}}$ of size $K_{\text{refined}}=130$.

Our loss consists in 3 terms, a orthogonality loss $L_{\text{orth}}(\*C_{\text{init}}) = \|\*C_{\text{init}}^\top \*C_{\text{init}} - I\|_2^2$, a consistency loss $L_{\text{consist}}(\*C_{\text{init}}, \*C_{\text{refined}}) = \|\*C_{\text{init}} - \*C_{\text{refined}}\|_2^2$, and a Laplacian bijectivity loss $L_{\text{lap}}(\*C_{\text{init}}) = \|\Delta \odot\*C_{\text{init}}\|_2^2$, where $\odot$ denotes element-wise product and $\Delta$ is obtained from~\cite{renStructuredRegularizationFunctional2019, caoUnsupervisedDeepMultishape2022}. More precisely, if $\lambda^{(1)},\lambda^{(2)}\in\RR^{K_\text{init}}$ denote the vector of eigenvalues of $S_1$ and $S_2$, then $\Delta$ is defined element-wise as
\begin{equation}
\begin{split}
    \Delta_{ij}^2 &= \left(\frac{\sqrt{\lambda^{(2)}_i}}{1+ \lambda^{(2)}_i} - \frac{\sqrt{\lambda^{(1)}_j}}{1+ \lambda^{(1)}_j}\right)^2\\
    &+ \left(\frac{1}{1+ \lambda^{(2)}_i} - \frac{1}{1+ \lambda^{(1)}_j}\right)^2 
\end{split}
\end{equation}

This is an extension of the standard Laplacian commutativity loss, which has been used in most existing implementations since GeoFMaps~\cite{donatiDeepGeometricFunctional2020}.

We do not enforce orthogonality of $\*C_{\text{refined}}$, since ZoomOut is proven to promote orthogonal functional maps~\cite{melziZoomOutSpectralUpsampling2019}. We notice that given a sound initialization, ZoomOut produces great results, which inspires us mostly to penalize  $\*C_{\text{init}}$. Furthermore, during the first iterations, initial functional maps produced by the network have no guarantee to be sound, and we therefore tune down the consistency loss $L_{\text{consist}}$ initially until the network converges towards good initialization. The consistency loss then provides meaningful guidance to the network. In practice, we increase the weight of this loss from $10^{-4}$ to $10^{-1}$ in 5 epochs using a multiplicative schedule.

The complete implementation is available at \url{https://github.com/RobinMagnet/SimplifiedFmapsLearning}.

\section{More Baselines \& Ablation}
\label{suppl:sec:baselines}

\begin{table}
\centering
{\def\arraystretch{1}\tabcolsep=0.5em
\setlength  \extrarowheight{1mm}

\begin{tabular}{cc}\toprule[0.8pt]
Method & SMAL  \\ \midrule[0.8pt]
ULRSSM~\cite{caoUnsupervisedLearningRobust2023} & 6.9   \\
ULRSSM (w/ fine-tune)~\cite{caoUnsupervisedLearningRobust2023} & \textbf{3.5} \\
AttentiveFMaps~\cite{liLearningMultiresolutionFunctional2022} & 5.4\\
ConsistentFMaps~\cite{sunSpatiallySpectrallyConsistent2023} & 5.4 \\ \midrule[0.5pt]
Ours (w/o Consistency) & 6.7 \\
Ours & 5.9\\
\bottomrule[0.8pt]
\end{tabular}}

\caption{Mean geodesic errors ($\times 100$) when training and testing on the SMAL dataset.  \label{tab:suppl:smal}}
\end{table}

We here present additional results on the standard baselines presented in the manuscript. In particular, some works~\cite{liLearningMultiresolutionFunctional2022,sunSpatiallySpectrallyConsistent2023} provided multiple versions of their algorithm. Furthermore, we display results from~\cite{caoUnsupervisedLearningRobust2023} using further test-time optimization. Note that this test-time optimization fine-tunes the network for each shape on the test set and should be applied to all other methods for fairness.
All these additional baselines can be found on~\Cref{tab:suppl:baselines}.

\begin{figure}
    \centering
    \includegraphics[width=.8\linewidth]{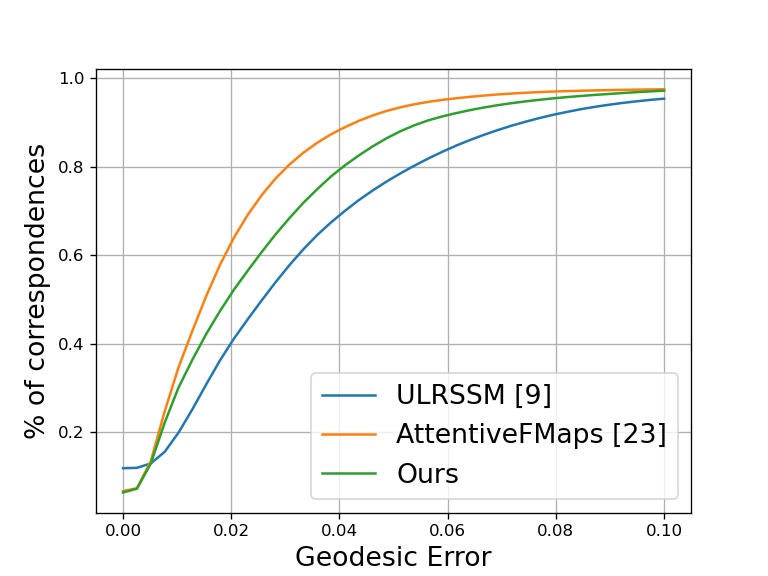}
    \caption{PCK curves on the SMAL dataset}
    \label{fig:suppl:pck curves}
\end{figure}

We additionally provide results on the SMAL dataset~\cite{zuffi3DMenagerieModeling2017}, where we additionally show the result of our pipeline without using the consistency loss (``w/o Consistency''), which serves as an ablation study similar to the one presented in~\cite{sunSpatiallySpectrallyConsistent2023}. However, in this ablation, we still use the ZoomOut algorithm at test-time, only the consistency loss was removed.
PCK curves for ULRSSM~\cite{caoUnsupervisedLearningRobust2023} and AttentiveFMaps~\cite{liLearningMultiresolutionFunctional2022} are also provided on \Cref{fig:suppl:pck curves}.

\section{ZoomOut Algorithm~\cite{melziZoomOutSpectralUpsampling2019}}

\begin{algorithm}
\caption{\label{algo:ZoomOut}The ZoomOut algorithm}
\begin{algorithmic}[1]
\Require Initial pointwise map $\Pi_{21}\in\{0,1\}^{n_2\times n_1}$ from $S_2$ to $S_1$, eigenvectors $\Phi_1\in\RR^{n\times k_1}$ and $\Phi_2^{n\times k_2}$ on each shape.
\vspace{.1cm}
\For{$k=k_{\text{init}}$ to $k_{\text{final}}$}
\State Compute $C_{12} = [\Phi_2]_{[:,:k]}^\dagger \Pi_{21} [\Phi_1]_{[:,:k]}$
\State Compute $\Pi_{21}=\text{NN}\big([\Phi_1]_{[:,:k]}C_{12}^\top, [\Phi_2]_{[:,:k]}\big)$
\EndFor
\State \textbf{Return} $C_{12}$, $\Pi_{21}$
\vspace{.2cm}
\end{algorithmic}
\end{algorithm}

The ZoomOut algorithm~\cite{melziZoomOutSpectralUpsampling2019} is a simple functional map refinement algorithm, which uses iterative conversions between functional and pointwise maps.

The algorithm is presented on \Cref{algo:ZoomOut}, where $\text{NN}$ denotes the nearest neighbor query between the rows of the two arguments.

\section{Adapting Scalable ZoomOut~\cite{magnetScalableEfficientFunctional2023}}

In~\cite{magnetScalableEfficientFunctional2023}, the authors present an approximation of the functional map for dense shapes using only sparse samples.

This approximation allows running the ZoomOut algorithm on a sparse subset of the vertices of both shapes, only using a complete high dimensional nearest neighbor query at the last step of the algorithm. This last step appears as the heaviest speed bottleneck of the algorithm as presented in~\cite{magnetScalableEfficientFunctional2023}.

When porting~\cite{magnetScalableEfficientFunctional2023} to GPU, this query makes the GPU run out of memory on very dense meshes, which we solve by using our scalable dense maps.

However, this algorithm adds a layer of approximation, which can potentially hinder the results. Furthermore, since it only uses values at sparse samples, the gradient can only propagates through these samples and not to the entire vertex-wise embeddings. In particular, it is not possible to use different samples each time the shape is used in training, as the preprocessing time is not negligible. This refrains us from using this adapted version within our learning framework.

\section{Dense Meshes}
\label{suppl:sec:Dense Meshes}

\begin{figure}
    \centering
    \includegraphics[width=.8\linewidth]{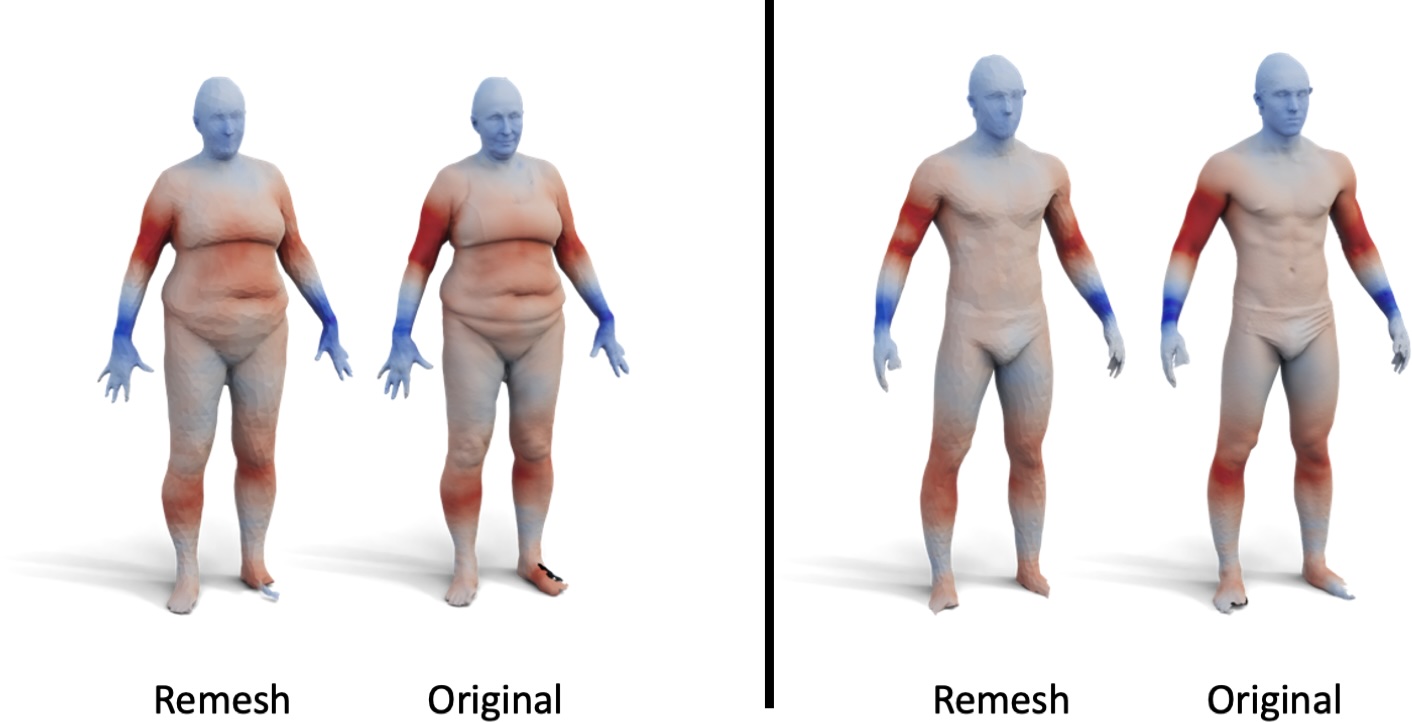}
    \caption{We leverage on the capacity of DiffusionNet~\cite{sharpDiffusionNetDiscretizationAgnostic2022a} to perform on various discretization of the same shape. Left and right are two shapes from the SHREC19 dataset~\cite{melziMatchingHumansDifferent2019}. We show on each shape features obtained on the remeshed and original version of the dataset.}
    \label{suppl:fig:dense features}
\end{figure}

\begin{figure}
    \centering
    \includegraphics[width=.4\textwidth]{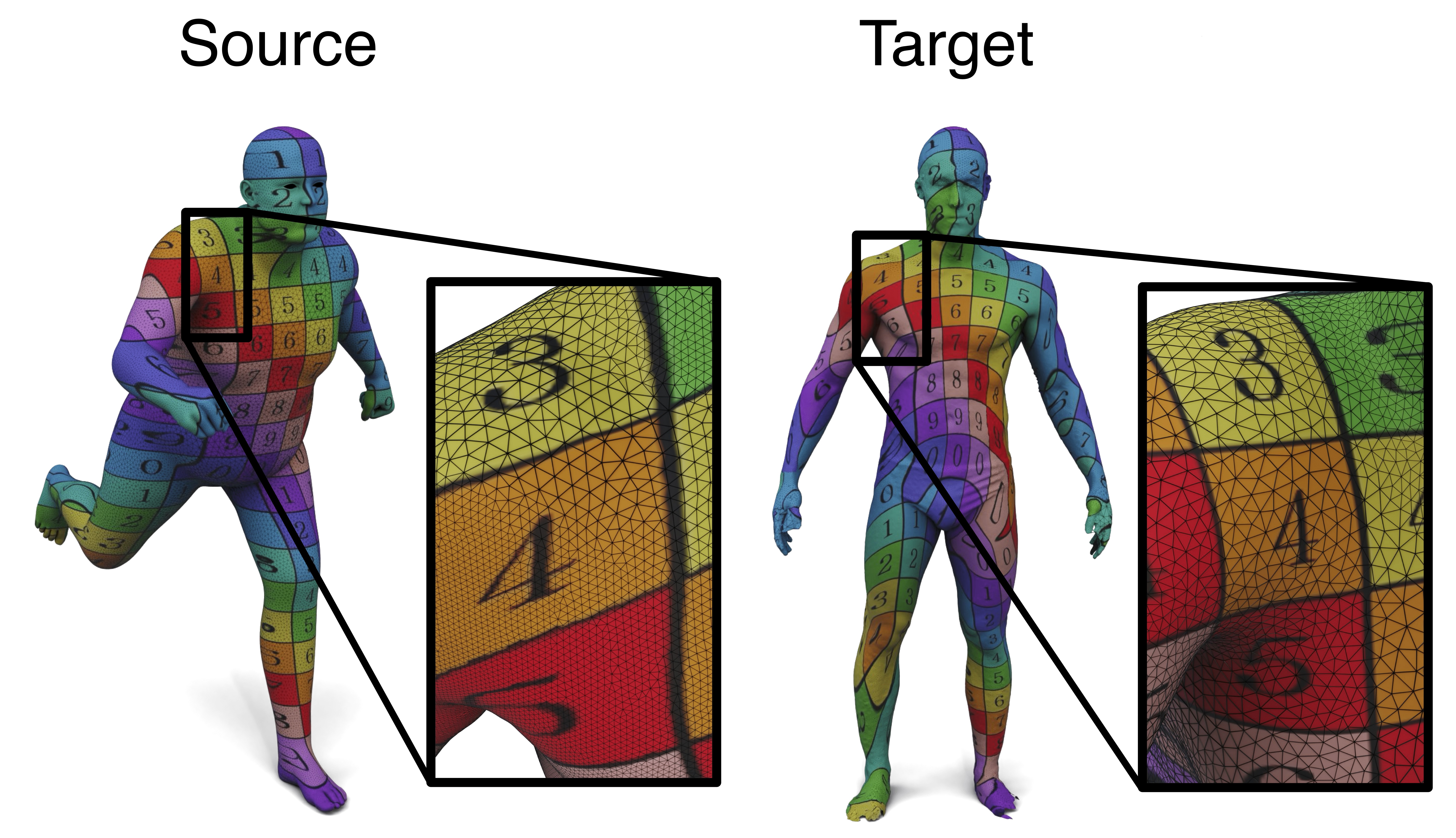}
    \caption{Example of texture transfer of our method on the SHREC19~\cite{melziMatchingHumansDifferent2019} dataset using our pipeline.}
    \label{fig:suppl:texture transfer}
\end{figure}

In this section, we provide more information on dense mesh processing using our pipeline, using meshes from the original version of the SHREC19 dataset~\cite{melziMatchingHumansDifferent2019}.

While DiffusionNet needs to store the eigenvectors of each shape of size $N\times K$ in memory, it is still able to compute features quickly for each shape. Due to its discretization-agnostic architecture, the features obtained on the dense and remeshed version are similar, as noted on \Cref{suppl:fig:dense features}, where each mesh contains $N=2\cdot 10^5$ vertices.
However, fitting a dense pointwise map would for this mesh require $10^7$ MiB of GPU memory, without even storing the gradient, which is infeasible in most cases.

In contrast, our scalable dense map can easily compute these maps. In particular, at test time when no gradient information is stored, our DifferentiableZoomOut has a negligible memory cost since intermediate maps don't need to be stored.

We show an example of texture transfer on another pair of this dataset in \Cref{fig:suppl:texture transfer}. Here, we used our network, trained on the standard remeshed~\cite{renContinuousOrientationpreservingCorrespondences2019} versions of the Faust~\cite{bogoFAUSTDatasetEvaluation2014} and Scape~\cite{anguelovSCAPEShapeCompletion2005} datasets, and evaluate at test time on shapes with around $10^5$ vertices. We transform the output functional map into a precise map~\cite{ezuzDeblurringDenoisingMaps2017}.
This demonstrates our pipeline can be trained on simple remeshed versions of datasets, but then used at test time on denser shapes without issues.

\end{document}